\documentclass[conference, a4paper]{IEEEtran}

\usepackage{graphicx}

\usepackage{epstopdf}
\usepackage{amsmath}
\usepackage{cite}
\usepackage{subcaption} % add in preamble
\usepackage{makecell}
\usepackage[bookmarks=false]{hyperref}  
\usepackage{dblfloatfix}  

\begin{document}

\title{Probabilistic Wildfire Susceptibility from Remote Sensing Using Random Forests and SHAP}

\author{\IEEEauthorblockN{1\textsuperscript{st} Udaya Bhasker Cheerala }
\IEEEauthorblockA{\textit{ Senior Consultant } \\
\textit{ Independent Researcher }\\
Bay Area, California, USA \\
0009-0000-9338-3327}
\and

\IEEEauthorblockN{2\textsuperscript{nd} Varun Teja Chirukuri }
\IEEEauthorblockA{\textit{ Software Developer } \\
\textit{ Columbia Sportswear }\\
Portland, Oregon, USA \\
0009-0000-5779-9931}
\and
\IEEEauthorblockN{3\textsuperscript{rd} Venkata Akhil Kumar Gummadi }
\IEEEauthorblockA{\textit{Graduate Researcher} \\
\textit{Indiana University}\\
Bloomington, Indiana, USA \\
0009-0003-1108-8044}
\and

\IEEEauthorblockN{\hspace{3.5cm}4\textsuperscript{th} Jintu Moni Bhuyan}
\IEEEauthorblockA{\hspace{3.5cm}\textit{Forestry and Ecology Department} \\
\hspace{3.5cm}\textit{Indian Institute of Remote Sensing}\\
\hspace{3.5cm}Dehradun, India\\
\hspace{3.5cm}0009-0002-1864-0957}
\and

\IEEEauthorblockN{5\textsuperscript{th} Praveen Damacharla }
\IEEEauthorblockA{\textit{ Research Scientist } \\
\textit{ KINETICAI INC }\\
The Woodlands, Texas, USA \\
praveen@kineticai.com \\ 
0000-0001-8058-7072}
}

\maketitle

\begin{abstract}

%(region transfer\footnote{\href{https://code.earthengine.google.com/bbcb6f5b4ea7b13a9bd3752f484a5dbc}{GEE Spatial Split Code}}) and temporal (2024--2025 split\footnote{\href{https://code.earthengine.google.com/a1d735b4685c2579b8d481074fbc03e0}{GEE Temporal Split Code}})

Wildfires pose a significant global threat to ecosystems worldwide, with California experiencing recurring fires due to various factors, including climate, topographical features, vegetation patterns, and human activities. This study aims to develop a comprehensive wildfire risk map for California by applying the random forest (RF) algorithm, augmented with Explainable Artificial Intelligence (XAI) through Shapley Additive exPlanations (SHAP), to interpret model predictions. Model performance was assessed using both spatial and temporal validation strategies. The RF model demonstrated strong predictive performance, achieving near-perfect discrimination for grasslands (AUC = 0.996) and forests (AUC = 0.997). Spatial cross-validation revealed moderate transferability, yielding ROC-AUC values of 0.6155 for forests and 0.5416 for grasslands. In contrast, temporal split validation showed enhanced generalization, especially for forests (ROC-AUC = 0.6615, PR-AUC = 0.8423). SHAP-based XAI analysis identified key ecosystem-specific drivers: soil organic carbon, tree cover, and Normalized Difference Vegetation Index (NDVI) emerged as the most influential in forests, whereas Land Surface Temperature (LST), elevation, and vegetation health indices were dominant in grasslands. District-level classification revealed that Central Valley and Northern Buttes districts had the highest concentration of high-risk grasslands, while Northern Buttes and North Coast Redwoods dominated forested high-risk areas. This RF-SHAP framework offers a robust, comprehensible, and adaptable method for assessing wildfire risks, enabling informed decisions and creating targeted strategies to mitigate dangers.

\end{abstract}


\section{Introduction}
\begin{figure}
    \centering
    \includegraphics[width=0.8\linewidth]{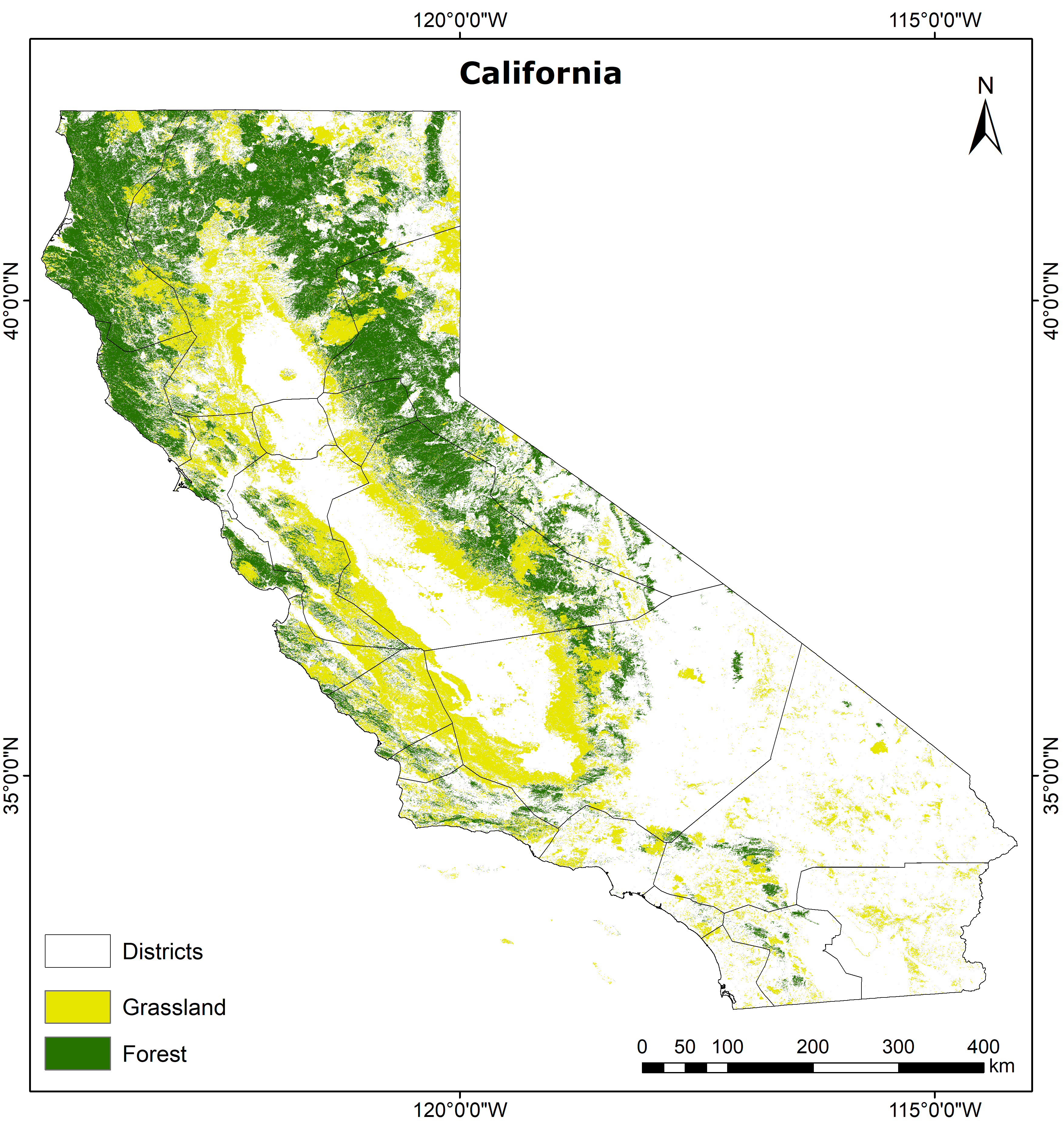}
    \caption{Study area in California showing forests, grasslands, and fire perimeters. (Source: NLCD) }
    \label{fig:placeholder}
\end{figure}
Wildfires are among the most widespread and severe threats to ecosystems worldwide, exerting profound environmental impacts \cite{Kurbanov2022}. Climate change has intensified wildfire risk, with rising temperatures and more frequent droughts increasing ecosystem vulnerability  \cite{Abatzoglou2019,Chuvieco2003,Ganteaume2013}. California is largely characterized as a “fuel-limited” ecosystem, with a fire regime spanning interior yellow pine, oak woodlands, grasslands, and mixed-conifer forests \cite{Keeley2021}. A critical factor influencing California wildfires is the availability of ignition sources \cite{Steel2015}. In the mountainous and desert regions of central California, frequent lightning strikes often trigger forest fires, whereas in the coastal regions, where lightning is rare, natural ignition sources are minimal. Nevertheless, coastal areas possess climatic conditions and fuel availability that remain highly conducive to the rapid spread of fires \cite{Steel2015}). Remote sensing imagery provides an effective approach for studying wildfires and the environmental factors influencing their occurrence, enabling large-scale monitoring, detailed assessment of vegetation and fuel conditions, and evaluation of the climatic drivers of forest fires \cite{MacDonald2020}.
\begin{figure*}[b]
    \centering
    \includegraphics[width=\textwidth]{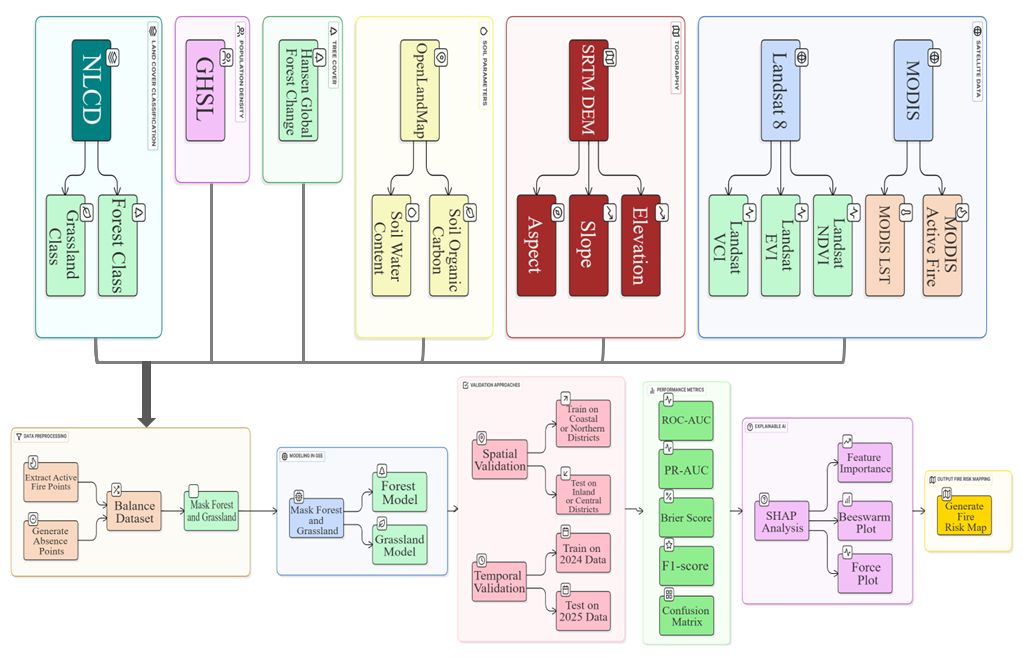}
    \caption{Proposed framework for remote sensing wildfire susceptibility using RF and SHAP.}
    \label{fig:methodology}
\end{figure*}
Wildfire risk prediction has been extensively studied, with existing models generally classified into three types: physics-based, semi-empirical, and empirical approaches \cite{Li2024}. These models often incorporate a multi-scale set of wildfire-related parameters, such as climatic factors, topographic attributes, and land cover and vegetation characteristics derived from satellite imagery \cite{Tonbul2024}. Geographic Information System (GIS) methods, coupled with multi-criteria decision-making (MCDM) approaches such as the analytic hierarchy process (AHP) and Fuzzy AHP, are frequently employed in spatial analysis studies \cite{Hodasova2025,Lamat2021,Ozcan2025,Pragya2023,Uthappa2025}. Advances in computational power and algorithm development have further facilitated the widespread adoption of machine learning (ML) algorithms for wildfire susceptibility mapping. Methods such as random forest (RF), support vector machines (SVM), gradient boosting machines (GBM), extreme gradient boosting (XGBoost), and artificial neural networks (ANN) have been increasingly utilized to capture complex patterns in environmental and climatic patterns, improving predictive accuracy for risk assessment, particularly, in estimating the spatial probability of fire occurrence. However, they are often regarded as “black-box” approaches, owing to their limited transparency in explaining the contribution of individual input variables to wildfire risk predictions at specific locations.

 Explainable Artificial Intelligence (XAI) addresses the interpretability challenges of ML models by enhancing their transparency and providing insights on internal decision-making processes\cite{Abdollahi2023,Tonbul2024}. XAI techniques enable visualization and interaction with model outputs, offering a clearer understanding of how predictions are derived. Incorporating interpretable techniques, such as SHapley Additive exPlanations (SHAP), can significantly enhance trust and comprehension of ML-driven wildfire risk assessments \cite{Abdollahi2023}.

Despite advances in wildfire modeling using GIS, MCDM, and ML approaches, several gaps remain. Many existing studies in California either lack spatially robust validation, rely on temporally constrained datasets, or fail to assess probabilistic model calibration. While ML models often achieve high predictive accuracy, they typically do not provide explicit measures of variable contributions, limiting their applicability for management and policy decisions. Moreover, few studies integrate XAI techniques with region-transfer and temporal validation, both of which are critical for generating reliable, generalizable, and operationally useful wildfire risk maps. Addressing these gaps is essential for providing decision-makers with transparent, interpretable, and robust wildfire susceptibility assessments across diverse Californian landscapes.
The present research focuses on developing a wildfire risk zonation model for California using ML methods, particularly the RF algorithm. In addition, XAI techniques, such as SHAP, are integrated to determine the most influential climatic, topographic, and anthropogenic factors driving wildfire occurrence and to quantify their relative importance within the predictive framework. Spatial and temporal cross-validation strategies are employed to ensure model robustness and generalizability across different regions and time periods. Overall, this study establishes an integrated ML–XAI framework to improve decision-making, enhance model interpretability, and pinpoint the most influential factors driving wildfire susceptibility.
\section{Study Area}
California, situated along the western coast of the United States, encompasses roughly 104.7 million acres, of which nearly 33 million acres are designated as forest land as in Fig.~\ref{fig:placeholder}. 
 %The state provides a unique setting of forest diversity and ecosystems, comprising redwood forests in the north, pine-dominated forests of the Sierra Nevada, oak woodlands, mixed conifer, mixed evergreen, red fir, and bigcone Douglas-fir (BCDF).
The state's climate is predominantly Mediterranean, characterized by cool, wet winters and hot, dry summers, with an intra-annual dry period of 3–6 months. These climatic gradients, combined with elevation patterns, play a pivotal role in shaping California's diverse fire regimes.
\section{Data and Methodology}
This study employed a range of geospatial datasets to model and map forest fire risk. MODIS data were used for active fire detection and land surface temperature measurements, while Landsat 8 imagery was utilized for vegetation indices, including NDVI, EVI, and VCI. Topographic variables—elevation, slope, and aspect—were derived from the SRTM DEM. Soil moisture and soil organic carbon were obtained from OpenLandMap, tree cover information from the Hansen Global Forest Change dataset, and population density from the Global Human Settlement Layer dataset. The National Land Cover Database (NLCD) was used to classify the study area into two primary land cover types: forested regions, comprising NLCD classes 41 (Deciduous Forest), 42 (Evergreen Forest), and 43 (Mixed Forest), and grassland areas, represented by class 71 (Grassland/Herbaceous). This classification provided a clearer basis for assessing fire dynamics and the factors influencing each fuel type.

Firstly, active fire occurrences from MODIS data were extracted for June–October 2025, corresponding to the region’s peak fire season as shown in Fig.~\ref{fig:methodology}. To create a balanced dataset for binary classification, an equal number of randomly generated absence points were included.

Subsequently, the RF model was implemented within the Google Earth Engine (GEE) platform to classify fire-prone and non-fire areas. To ensure robustness, both spatial transfer (region-to-region generalization) and temporal split (2024 vs. 2025) validation strategies were employed, allowing assessment of the model’s predictive stability across space and time. Key evaluation metrics included Positive Rate (fires detected), ROC-AUC, PR-AUC, Brier Score, Overall Accuracy, Precision, Recall, and F1-score, all computed from the test dataset’s confusion matrix. To enhance interpretability, SHAP-based XAI was employed to quantify the contributions of predictor variables—such as vegetation, temperature, and slope—to wildfire risk, thereby providing actionable insights for decision-makers.

\section{Results and Analysis}
\subsection{Model Training and Validation Performance}
The RF model exhibited excellent performance in predicting fire occurrences across both  forest and grassland ecosystems. For forest, the model performed exceptionally well, yielding an AUC of $0.997$ and an overall accuracy of $94.9\%$, with an F1-score of $0.963$. The recall was notably high ($0.997$), suggesting that the model effectively identified fire events with minimal false negatives, although the precision ($0.930$) was slightly lower, reflecting a few mis-classifications of non-fire pixels. For grassland, the model achieved a very high discriminatory power with an AUC of $0.996$, an accuracy of $96.3\%$, and an F1-score of $0.968$. The precision and recall were $0.951$ and $0.985$, respectively, indicating strong reliability in detecting both fire and non-fire events as shown in Table \ref{tab:rf_performance}. 

\begin{table}[htbp]
\centering
\caption{RF model performance for forest and grassland ecosystems.}
\label{tab:rf_performance}
\resizebox{\linewidth}{!}{%
\begin{tabular}{l c c c c c}
\hline
Ecosystem & AUC & Accuracy & Precision & Recall & F1-score \\ 
\hline
Forest & 0.997 & 0.949 & 0.930 & 0.997 & 0.963 \\ 
Grassland & 0.996 & 0.963 & 0.951 & 0.985 & 0.968 \\ 
\hline
\end{tabular}%
}
\end{table}

% ===== Confusion Matrix =====
\begin{figure}[!ht]
    \centering
    \begin{subfigure}[b]{0.48\linewidth}
        \centering
        \includegraphics[width=\linewidth]{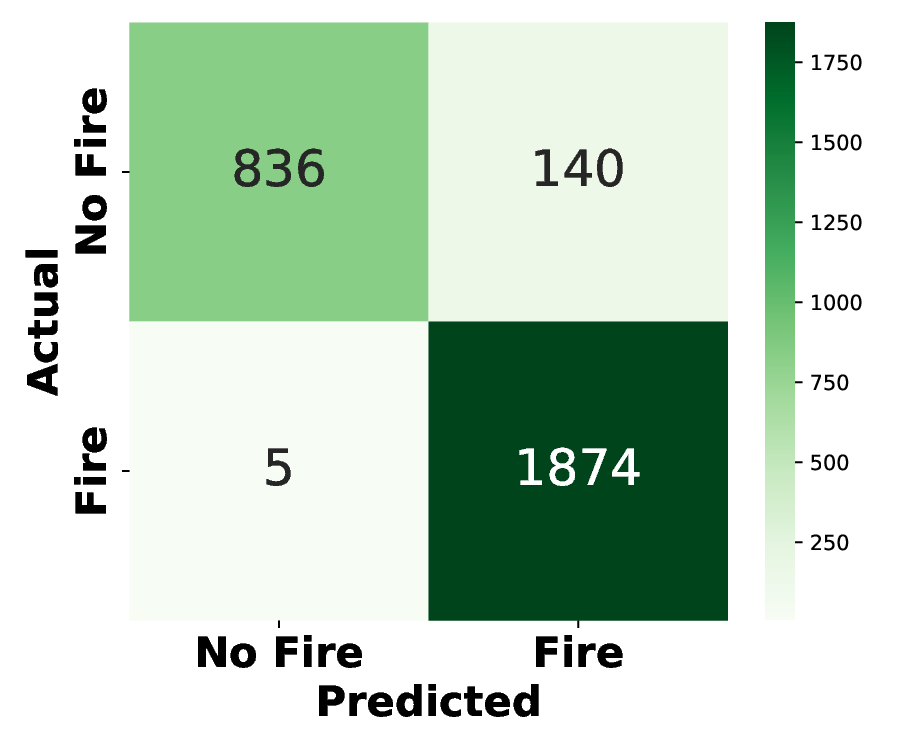}
        \caption{Forest}
        \label{fig:cm_forest}
    \end{subfigure}
    \hfill
    \begin{subfigure}[b]{0.48\linewidth}
        \centering
        \includegraphics[width=\linewidth]{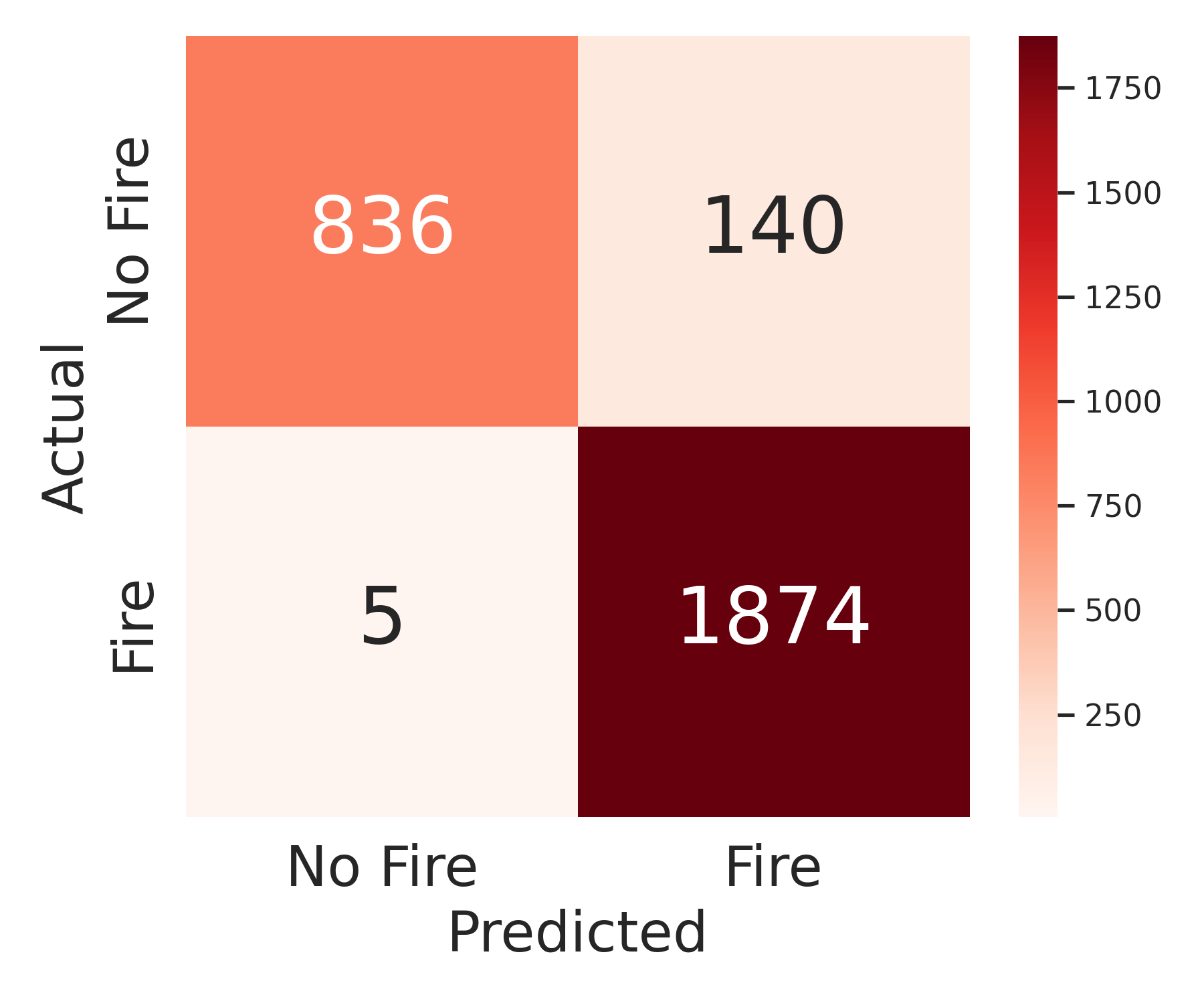}
        \caption{Grassland}
        \label{fig:cm_grassland}
    \end{subfigure}
    \caption{Confusion matrix of the RF classifier for forest and grassland datasets.}
    \label{fig:confusion_matrices}
\end{figure}

% ===== ROC Curves =====
\begin{figure}[!ht]
    \centering
    \begin{subfigure}[b]{0.48\linewidth}
        \centering
        \includegraphics[width=\linewidth]{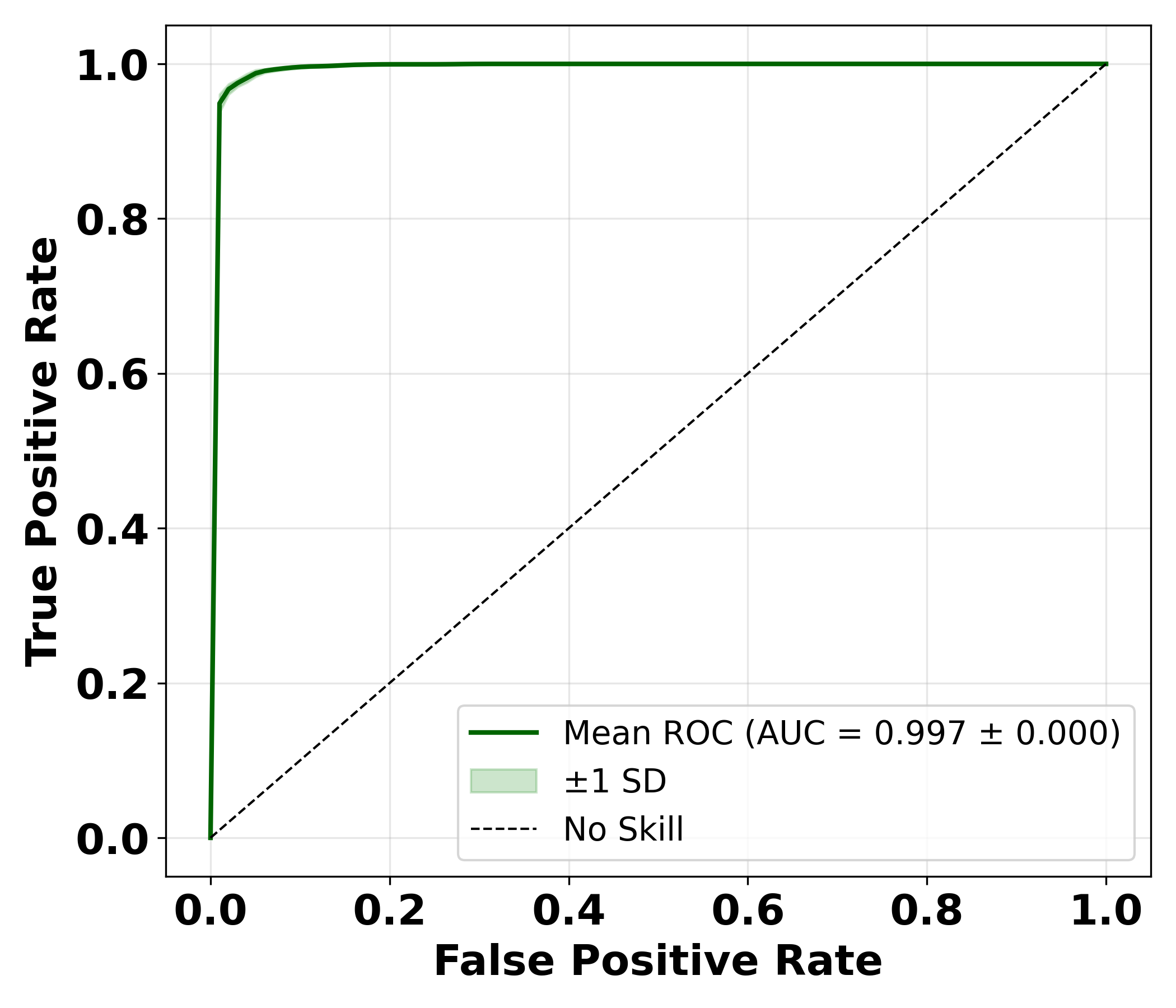}
        \caption{Forest}
        \label{fig:roc_forest}
    \end{subfigure}
    \hfill
    \begin{subfigure}[b]{0.48\linewidth}
        \centering
        \includegraphics[width=\linewidth]{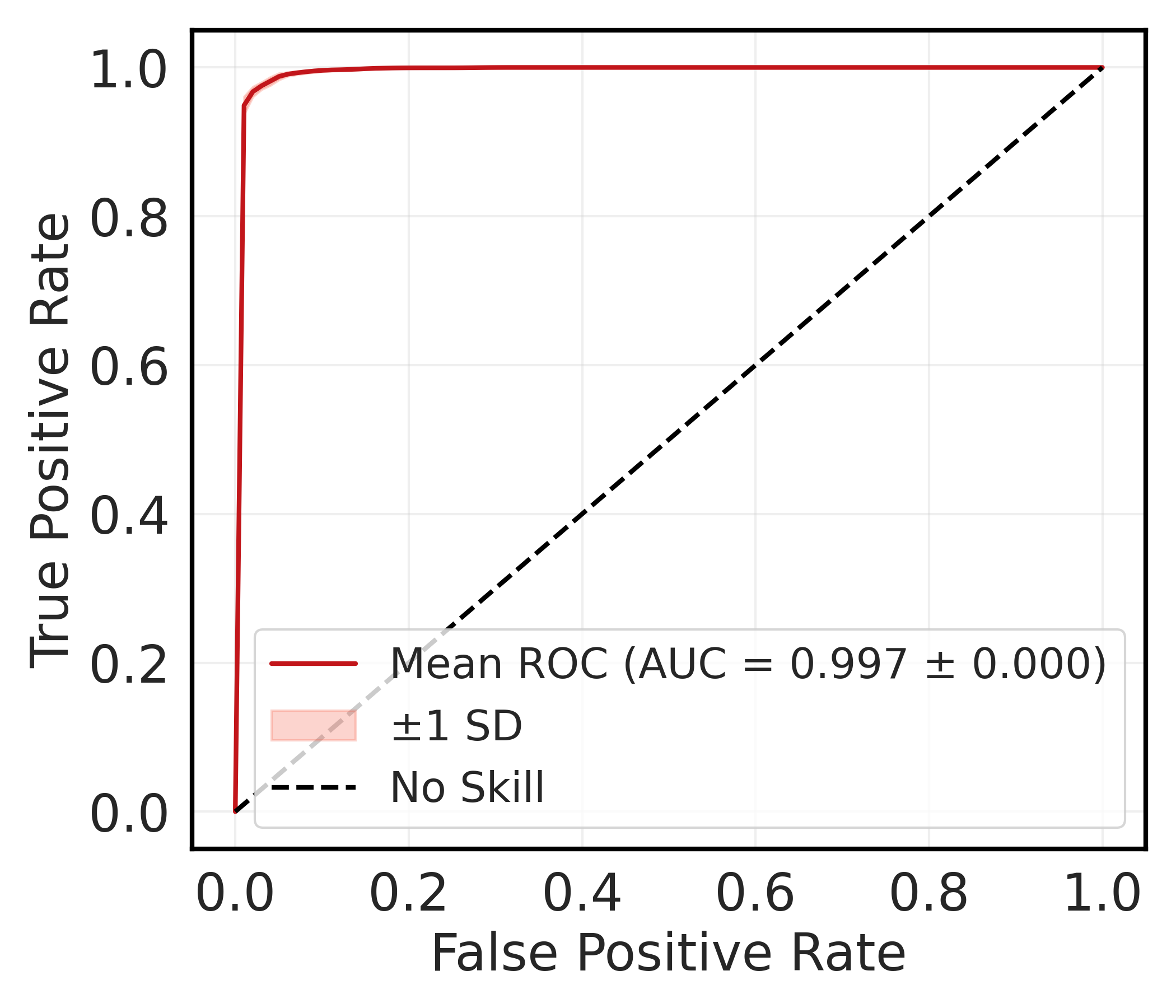}
        \caption{Grassland}
        \label{fig:roc_grassland}
    \end{subfigure}
    \caption{ROC curves with bootstrapped confidence intervals for the RF classifier applied to forest and grassland datasets.}
    \label{fig:roc_curves}
\end{figure}

% ===== Precision-Recall Curves =====
\begin{figure}[!ht]
    \centering
    \begin{subfigure}[b]{0.48\linewidth}
        \centering
        \includegraphics[width=\linewidth]{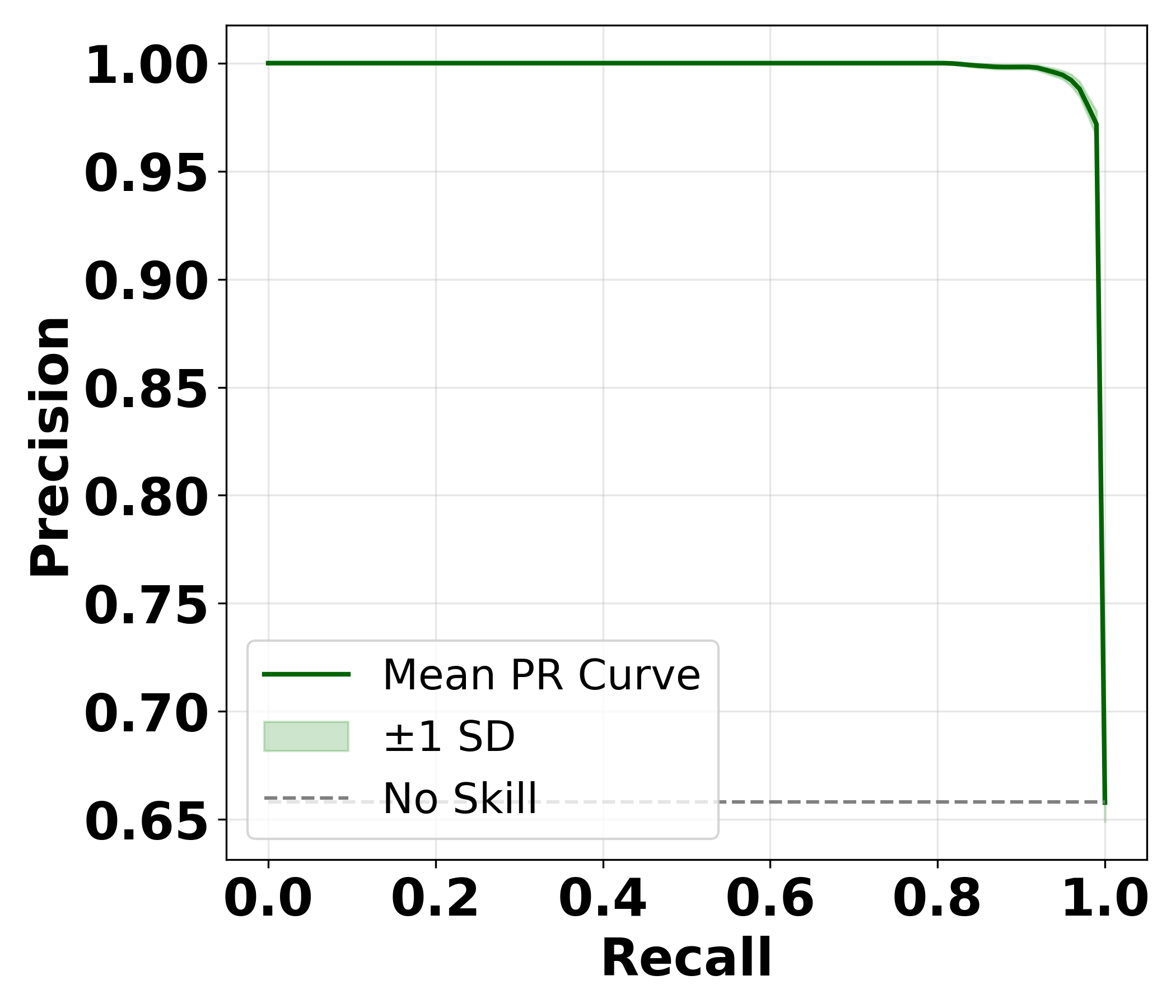}
        \caption{Forest}
        \label{fig:pr_forest}
    \end{subfigure}
    \hfill
    \begin{subfigure}[b]{0.48\linewidth}
        \centering
        \includegraphics[width=\linewidth]{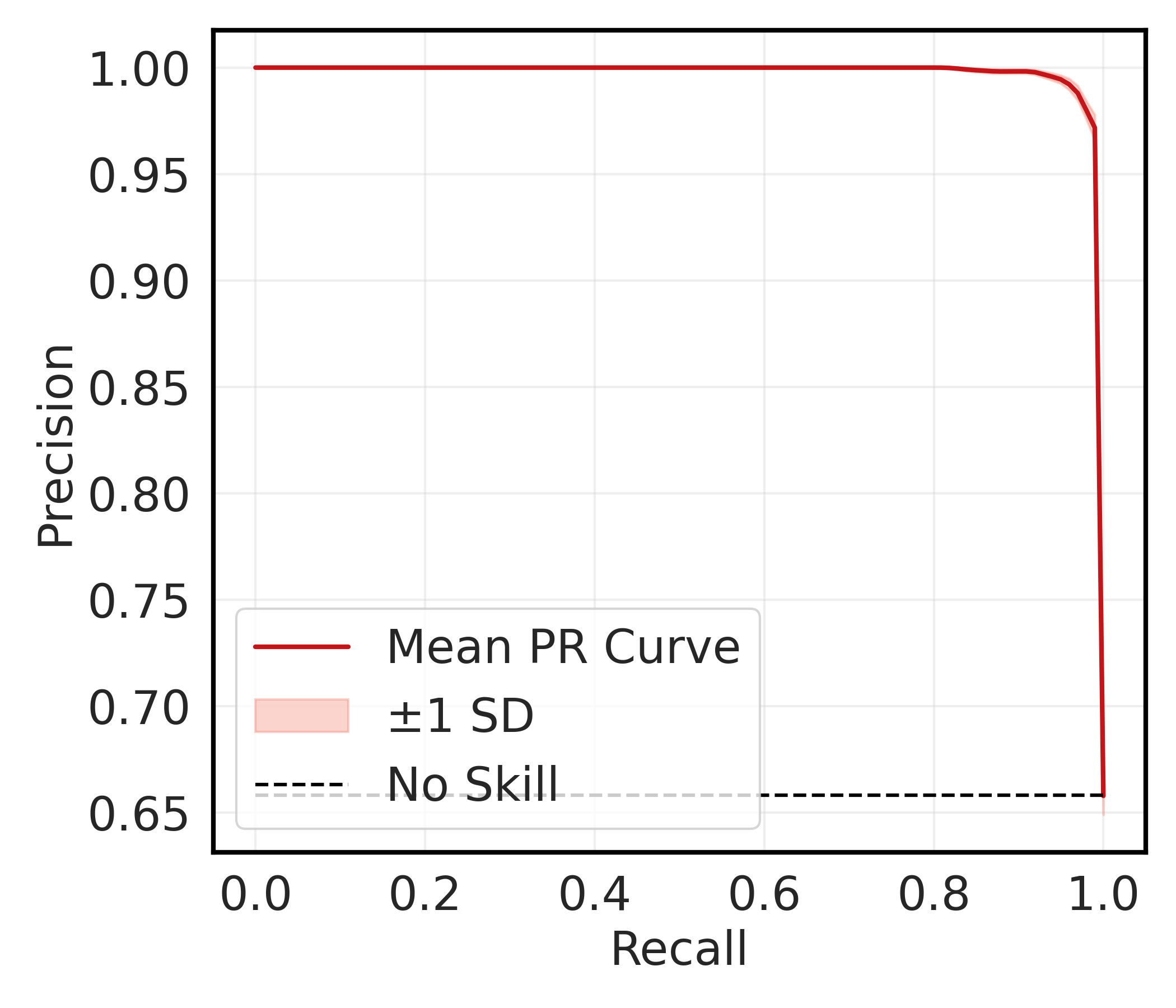}
        \caption{Grassland}
        \label{fig:pr_grassland}
    \end{subfigure}
    \caption{PR curves with confidence intervals for the RF classifier applied to forest and grassland datasets.}
    \label{fig:pr_curves}
\end{figure}

For the forest ecosystem, the RF classifier correctly identified 1874 fire events and 836 non-fire events, with only 5 missed fire events (false negatives) and 140 false alarms (false positives) as shown in Fig. \ref{fig:confusion_matrices}(a). For grasslands, the confusion matrix in Fig. \ref{fig:confusion_matrices}(b) shows 1900 correctly predicted fire events and 1399 correctly predicted non-fire events, alongside 29 missed fire events and 98 false positives. The corresponding ROC curves showed an AUC of 0.997 ± 0.000 (forest) and AUC of 0.996 ± 0.001 (grassland), reflecting near-perfect discrimination between fire and non-fire pixels  as in Fig. \ref{fig:roc_curves}. Similarly, the Precision-Recall (PR) curve in Fig. \ref{fig:pr_curves} remained close to 1.0 across thresholds. However, compared to the forest case scenario, the grassland results showed a modest increase in false negatives. The PR curve confirmed this balance, with consistently high precision and recall values.
 
 %The confusion matrix (Figure \ref{fig:confusion_matrices}) shows 1900 correctly predicted fire events and 1399 correctly predicted no-fire events, alongside 29 missed fire events and 98 false positives for grassland. The AUC of 0.996 ± 0.001 (Figure \ref{fig:confusion_matrices})
%However, compared to the forest case, the grassland results show a modest increase in false negatives. The PR curve (Figure \ref{fig:pr_curves}) confirmed this balance, with consistently high precision and recall values.

\subsection{Cross Validation}

 %\subsubsection{Region Transfer Validation}

The fire occurrence model was trained using an RF classifier on coastal and northern California districts and was evaluated through spatial cross-validation on the independent inland, Sierra, and central districts. 
% Forest vs Grassland Figures Side by Side
% Forest vs Grassland Figures Side by Side
% ROC Curve
\begin{figure}[htbp]
    \centering
    \begin{subfigure}[b]{0.49\linewidth}
        \centering
        \includegraphics[width=\linewidth]{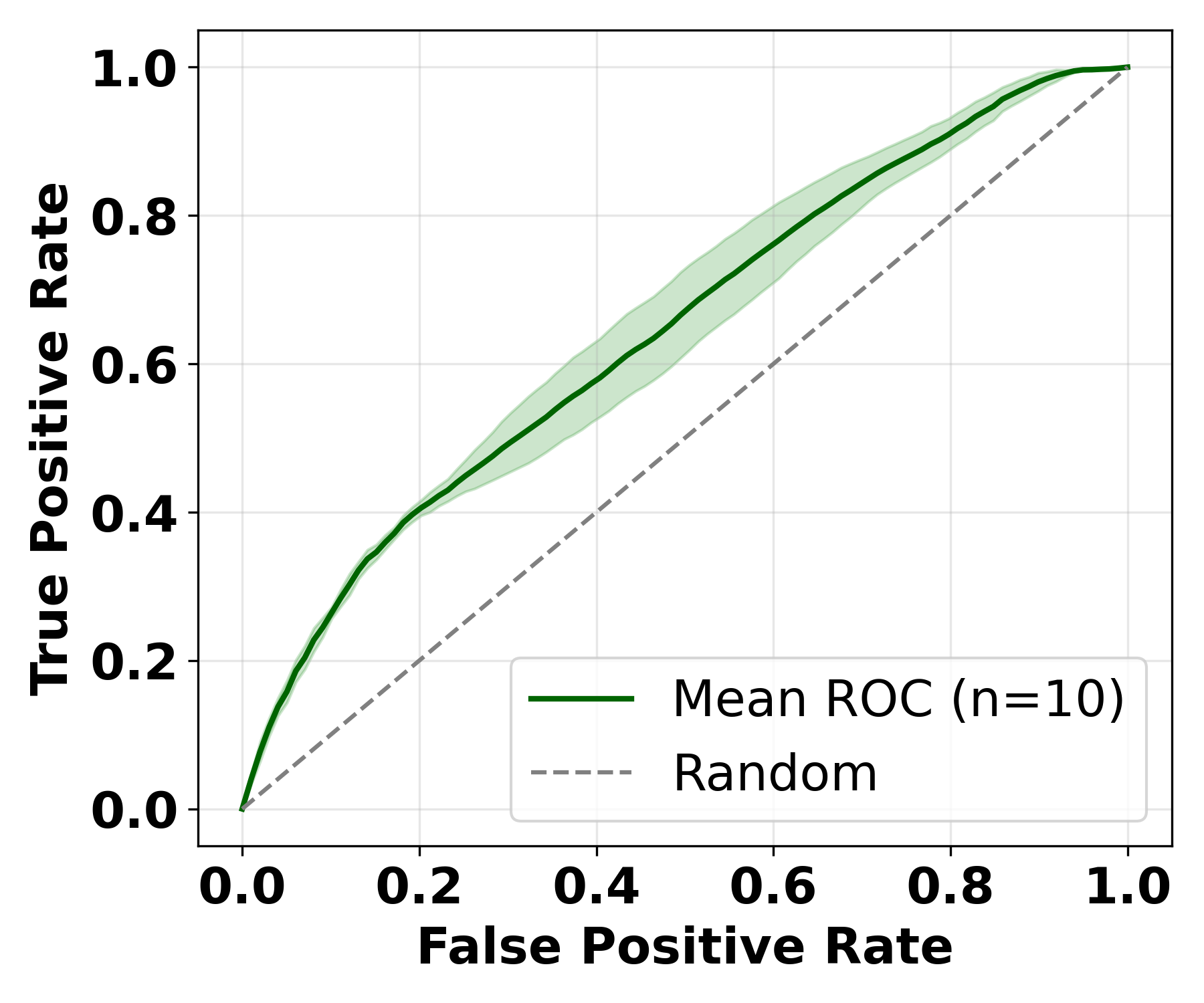}
        \caption{Forest}
        \label{fig:F_roc_curve}
    \end{subfigure}
    \hfill
    \begin{subfigure}[b]{0.49\linewidth}
        \centering
        \includegraphics[width=\linewidth]{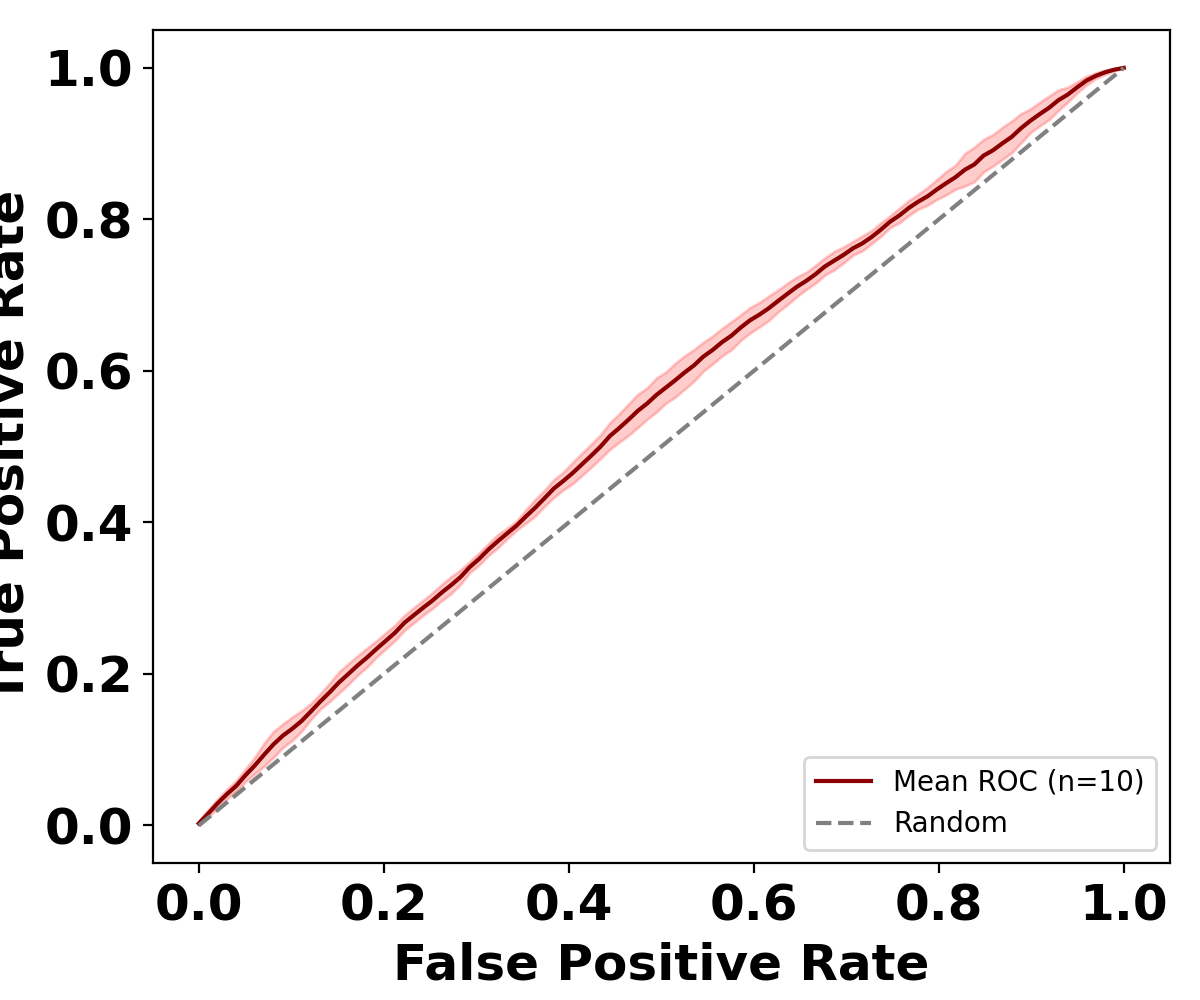}
        \caption{Grassland}
        \label{fig:G_roc_curve}
    \end{subfigure}
    \caption{ROC curves with mean and standard deviation for forest and grassland regions.}
    \label{fig:ROC_curve_F_G}
\end{figure}

\begin{figure}[htbp]
    \centering
    % PR Curve
    \begin{subfigure}[b]{0.49\linewidth}
        \centering
        \includegraphics[width=\linewidth]{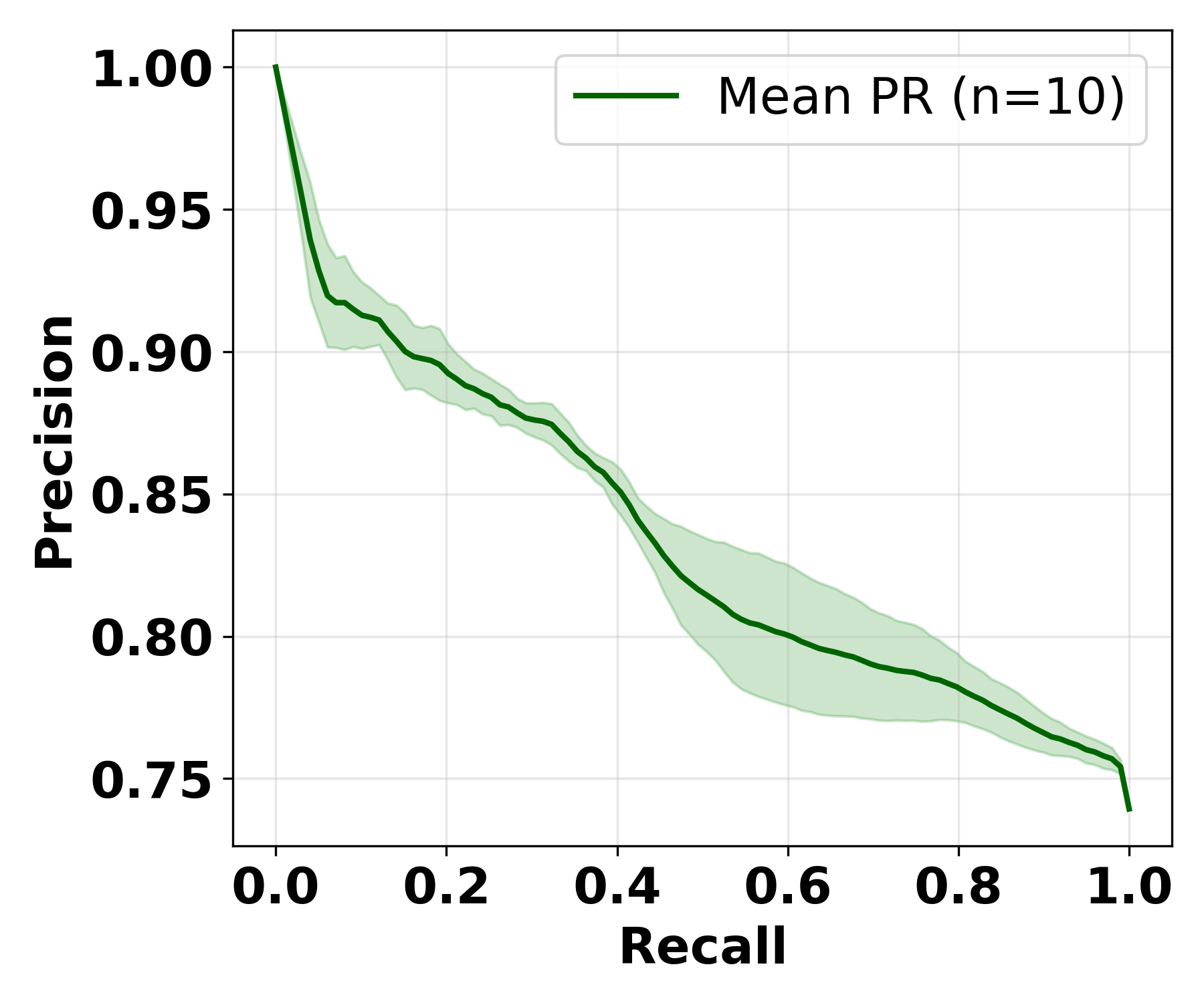}
        \caption{Forest}
        \label{fig:F_pr_curve}
    \end{subfigure}
    \hfill
    \begin{subfigure}[b]{0.49\linewidth}
        \centering
        \includegraphics[width=\linewidth]{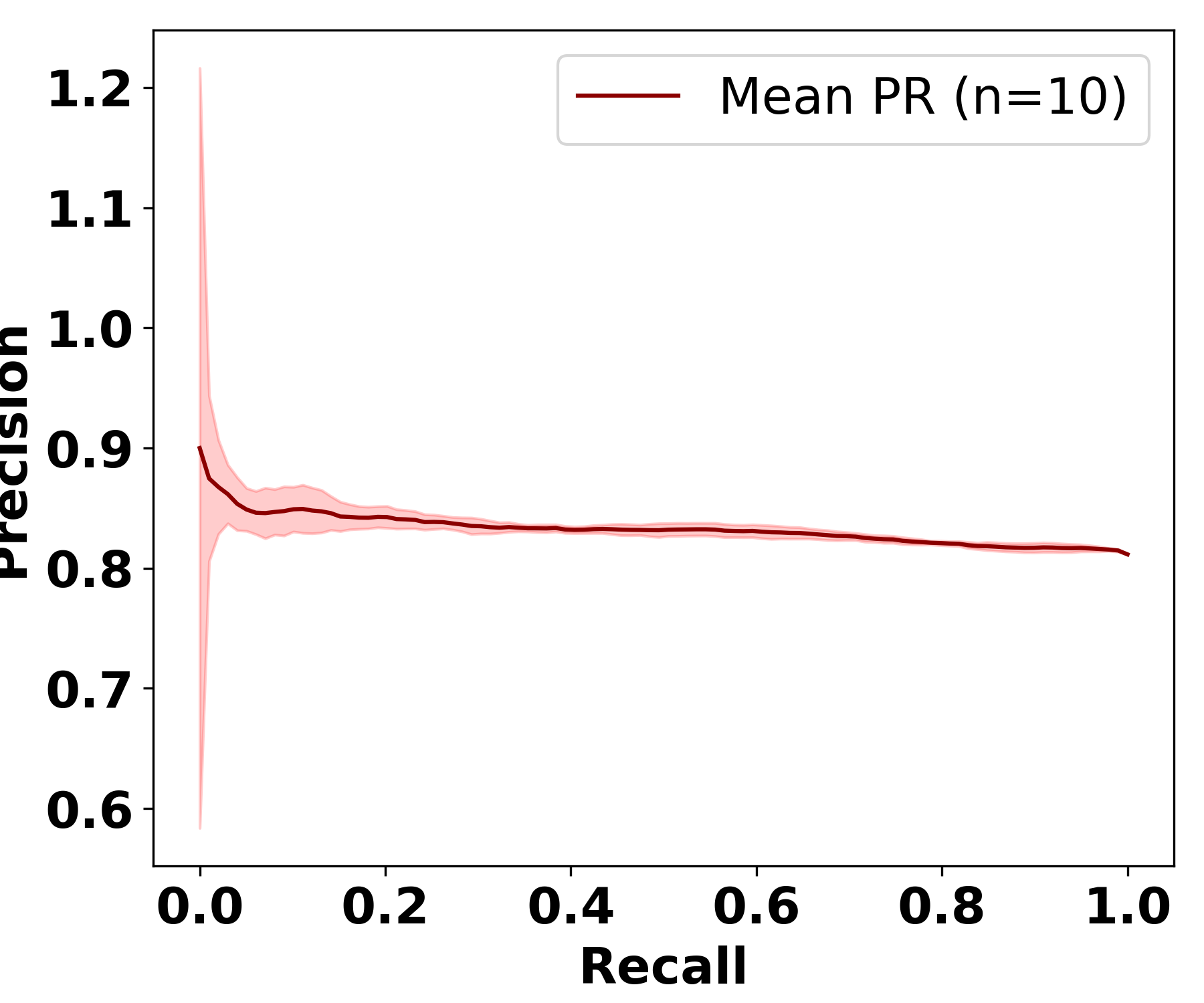}
        \caption{Grassland}
        \label{fig:G_pr_curve}
    \end{subfigure}
    \caption{PR curve with mean and standard deviation for forest and grassland regions.}
    \label{fig:PR_curve_F_G}
\end{figure}

% Reliability Curve
\begin{figure}[htbp]
    \centering
    \begin{subfigure}[b]{0.49\linewidth}
        \centering
        \includegraphics[width=\linewidth]{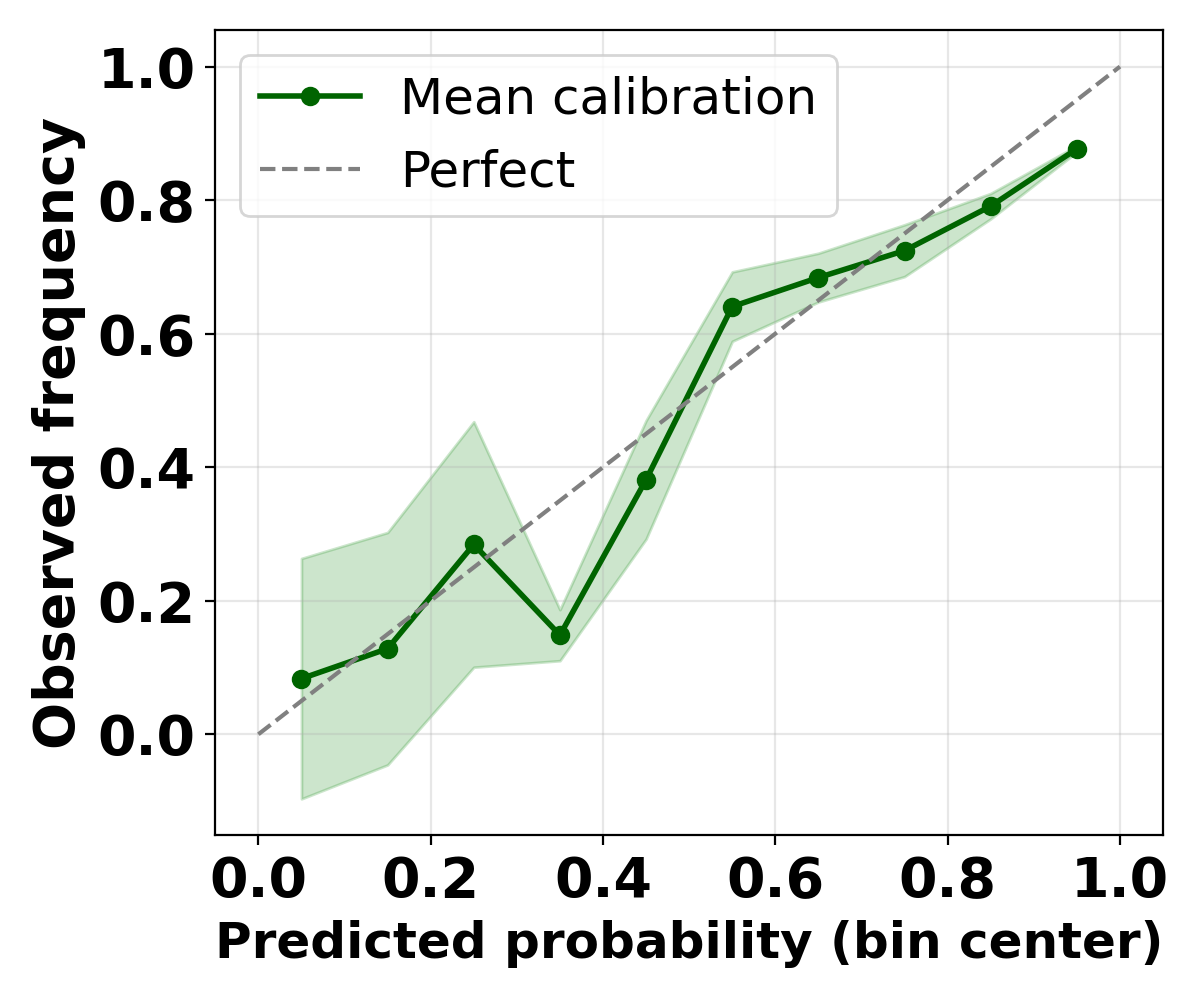}
        \caption{Forest}
        \label{fig:F_reliability}
    \end{subfigure}
    \hfill
    \begin{subfigure}[b]{0.49\linewidth}
        \centering
        \includegraphics[width=\linewidth]{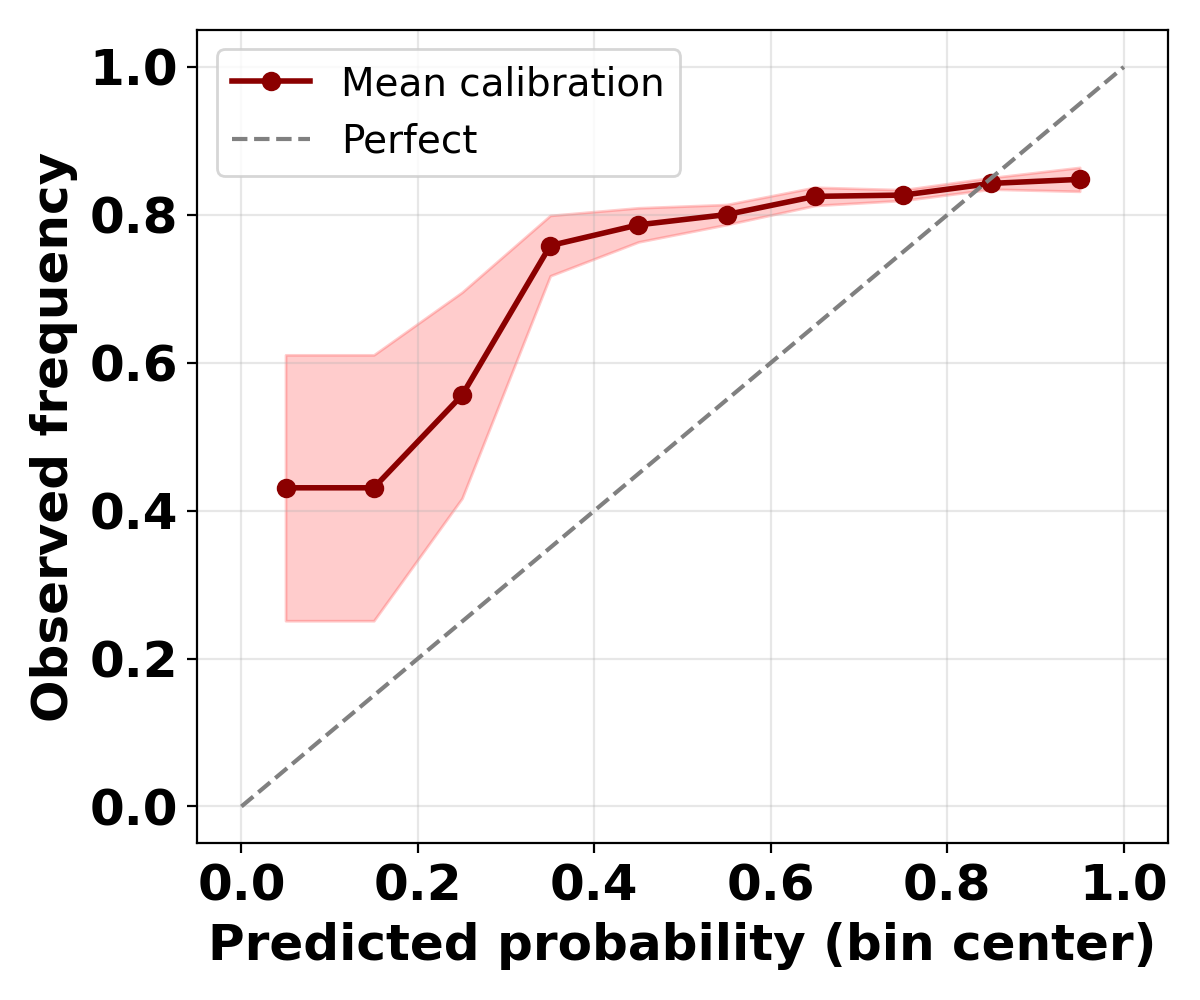}
        \caption{Grassland}
        \label{fig:G_reliability}
    \end{subfigure}
    \caption{Reliability diagrams showing mean and standard deviation for forest and grassland regions.}
    \label{fig:Reliability_curve_F_G}
\end{figure}

% Top-K Curve
\begin{figure}[htbp]
    \centering
    \begin{subfigure}[b]{0.49\linewidth}
        \centering
        \includegraphics[width=\linewidth]{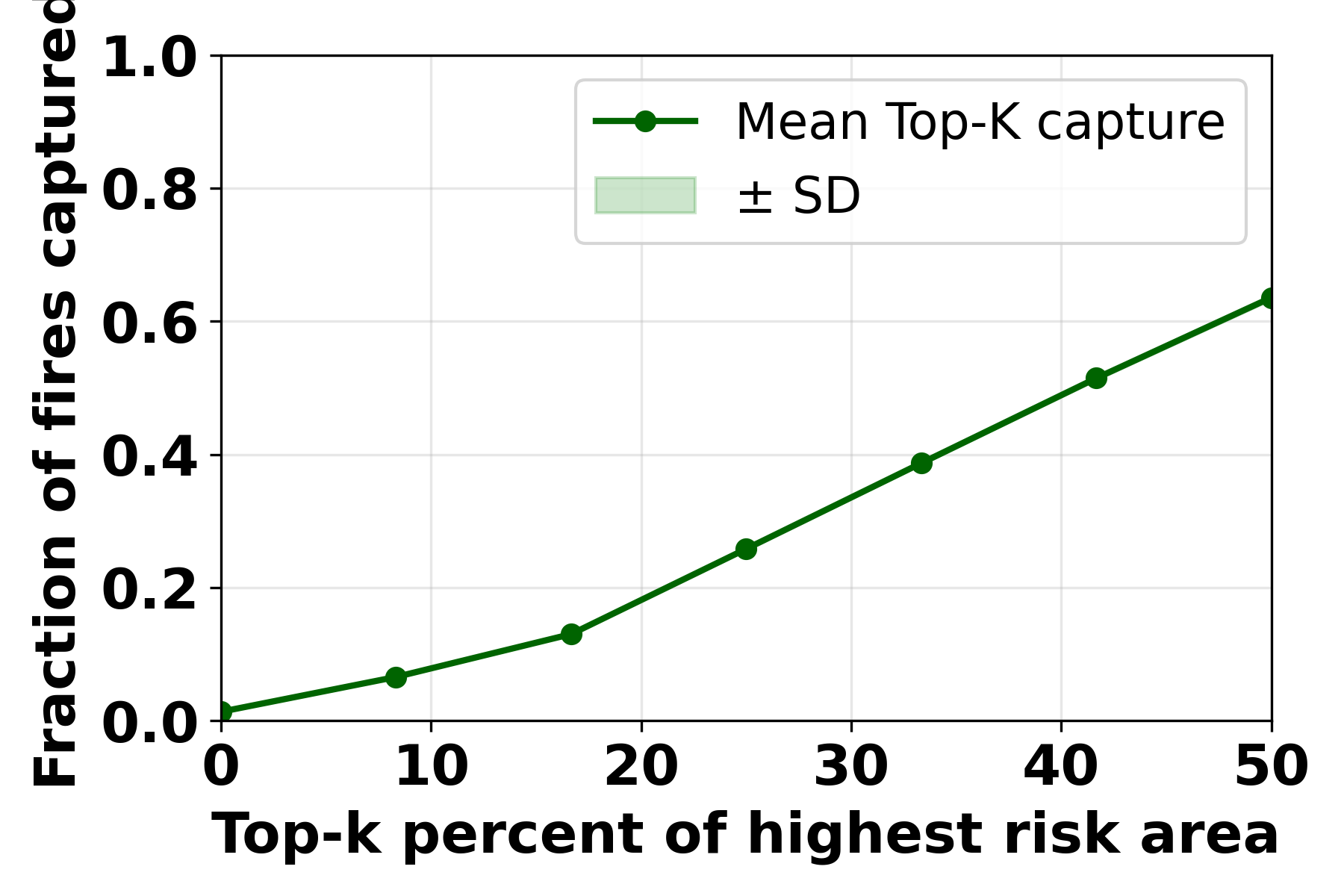}
        \caption{Forest}
        \label{fig:F_topk_curve}
    \end{subfigure}
    \hfill
    \begin{subfigure}[b]{0.49\linewidth}
        \centering
        \includegraphics[width=\linewidth]{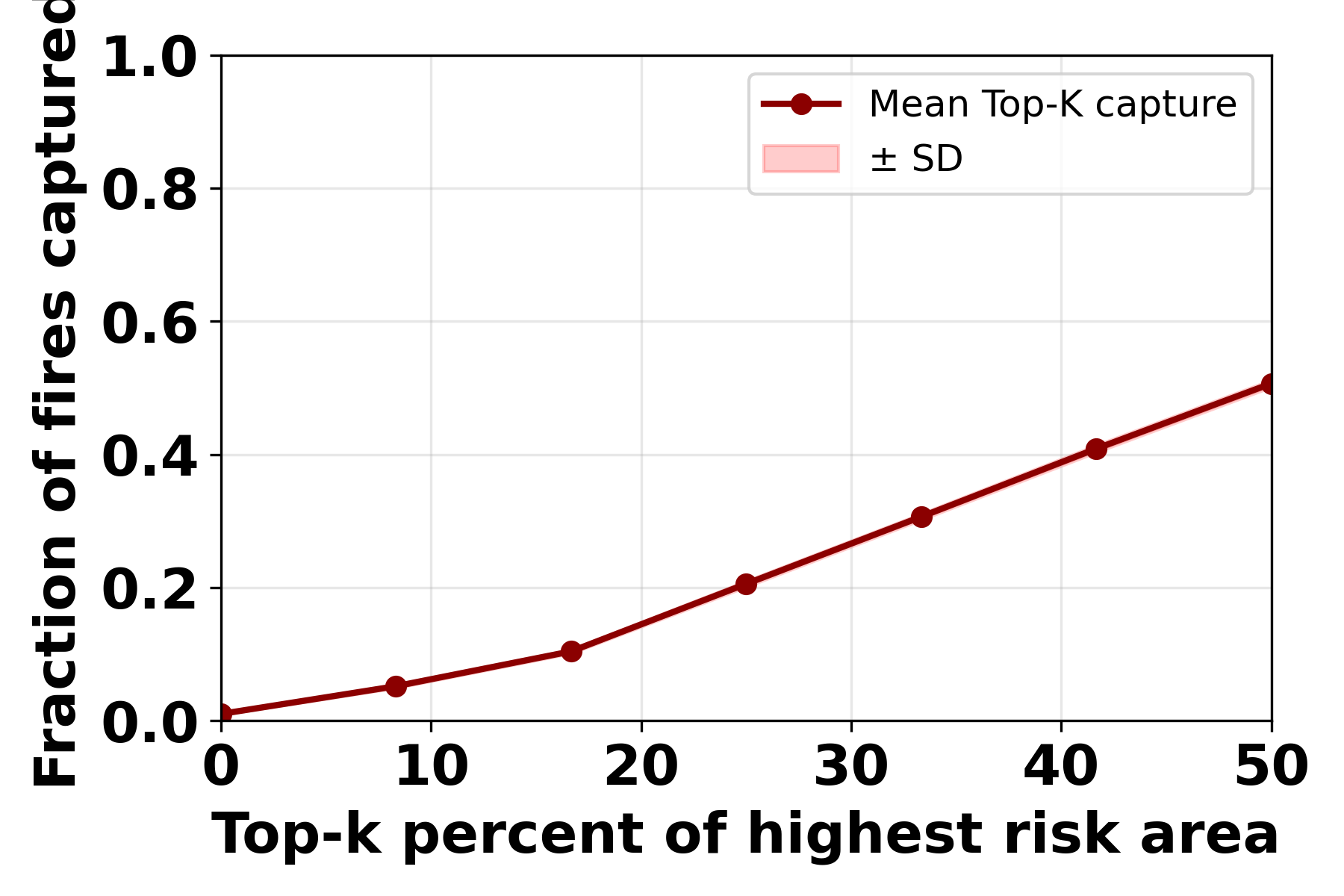}
        \caption{Grassland}
        \label{fig:G_topk_curve}
    \end{subfigure}
    \caption{Top-$k$ capture curves for forest and grassland regions.}
    \label{fig:TopK_curve_F_G}
\end{figure}

The classifier showed a mixed but interpretable performance profile. 
Discrimination measured by ROC-AUC was moderate for the forest stratum 
(mean ROC-AUC $\approx 0.658  \pm  0.0106$) and only marginally better than random 
for grassland (mean ROC-AUC $\approx 0.521 \pm 0.0179$), indicating limited ability 
to separate positives from negatives across all thresholds as shown in Fig. \ref{fig:ROC_curve_F_G}. 

By contrast, PR performance was strong 
(forest PR-AUC $\approx 0.842 \pm 0.0082$, grassland PR-AUC $\approx 0.824 \pm 0.0102$). 
This apparent discrepancy (moderate ROC but high PR) is explained by the high prevalence  of the positive class in both strata  (positive rate $\approx 0.740$ in forest, $0.810$ in grassland): when positives dominate,  PR curves and PR-AUC can remain high even if the model does not sharply separate classes as shown in Fig. \ref{fig:PR_curve_F_G}. 

At the $0.5$ threshold the forest model attained a very high recall  ($\sim 0.965$) with reasonable precision ($\sim 0.756$), producing a high  $F_{1}$ score ($\approx 0.848$); grassland had higher precision ($\sim 0.81$)  but lower recall ($\sim 0.779$), and correspondingly lower $F_{1}$ score ($\approx 0.794$). 

The top-$k$ capture results further showed modest enrichment: the top $50\%$ of ranked 
predictions capture $\sim 55.7\%$ of forest positives and $\sim 50.6\%$ of grassland 
positives ($\text{mean\_frac}$) Fig. \ref{fig:TopK_curve_F_G}.

\begin{figure}[htbp]
    \centering
    % Forest ROC
    \begin{subfigure}[b]{0.49\linewidth}
        \centering
        \includegraphics[width=\linewidth]{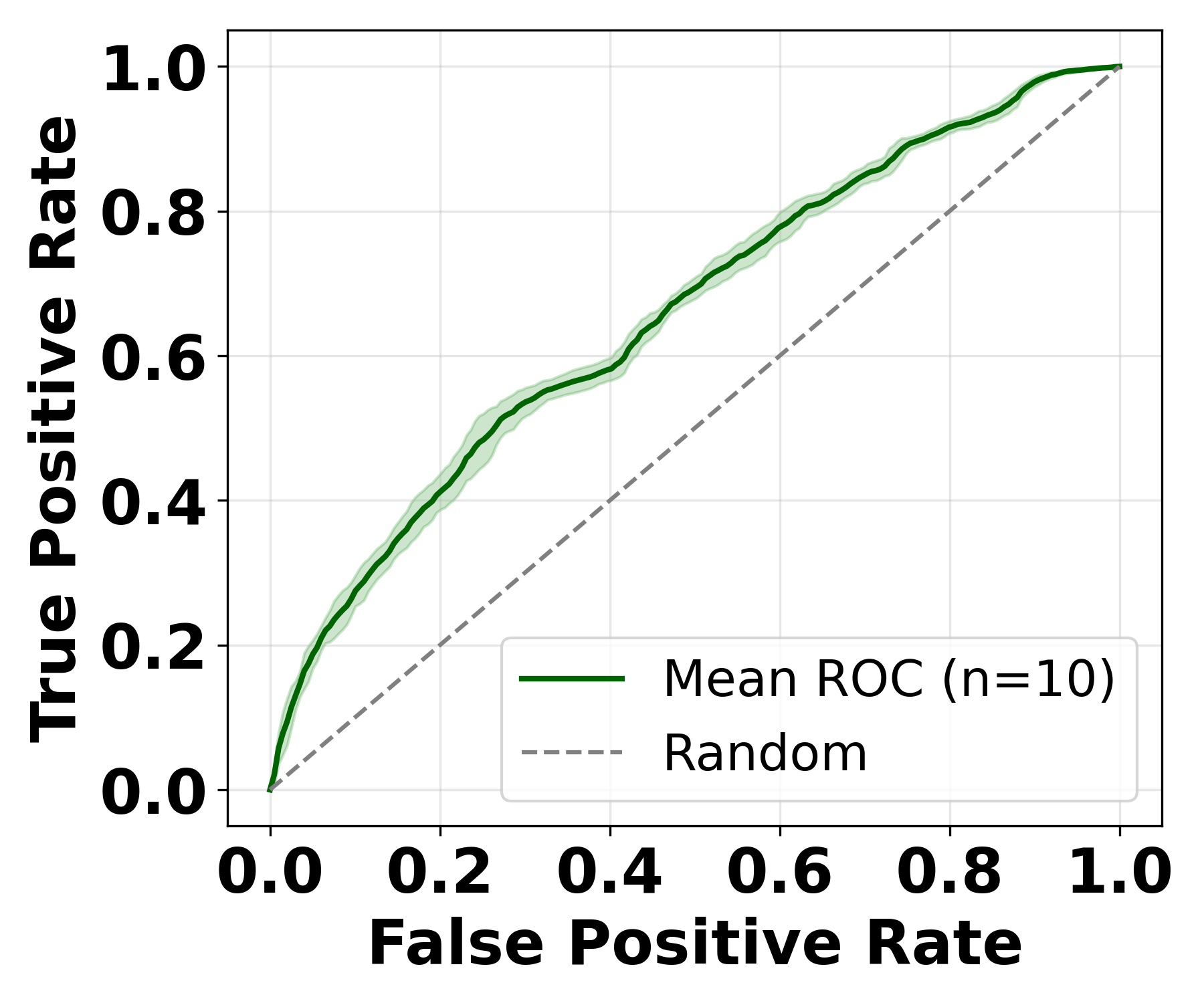}
        \caption{Forest}
        \label{fig:TS_F_roc_curve}
    \end{subfigure}
    \hfill
    % Grassland ROC
    \begin{subfigure}[b]{0.49\linewidth}
        \centering
        \includegraphics[width=\linewidth]{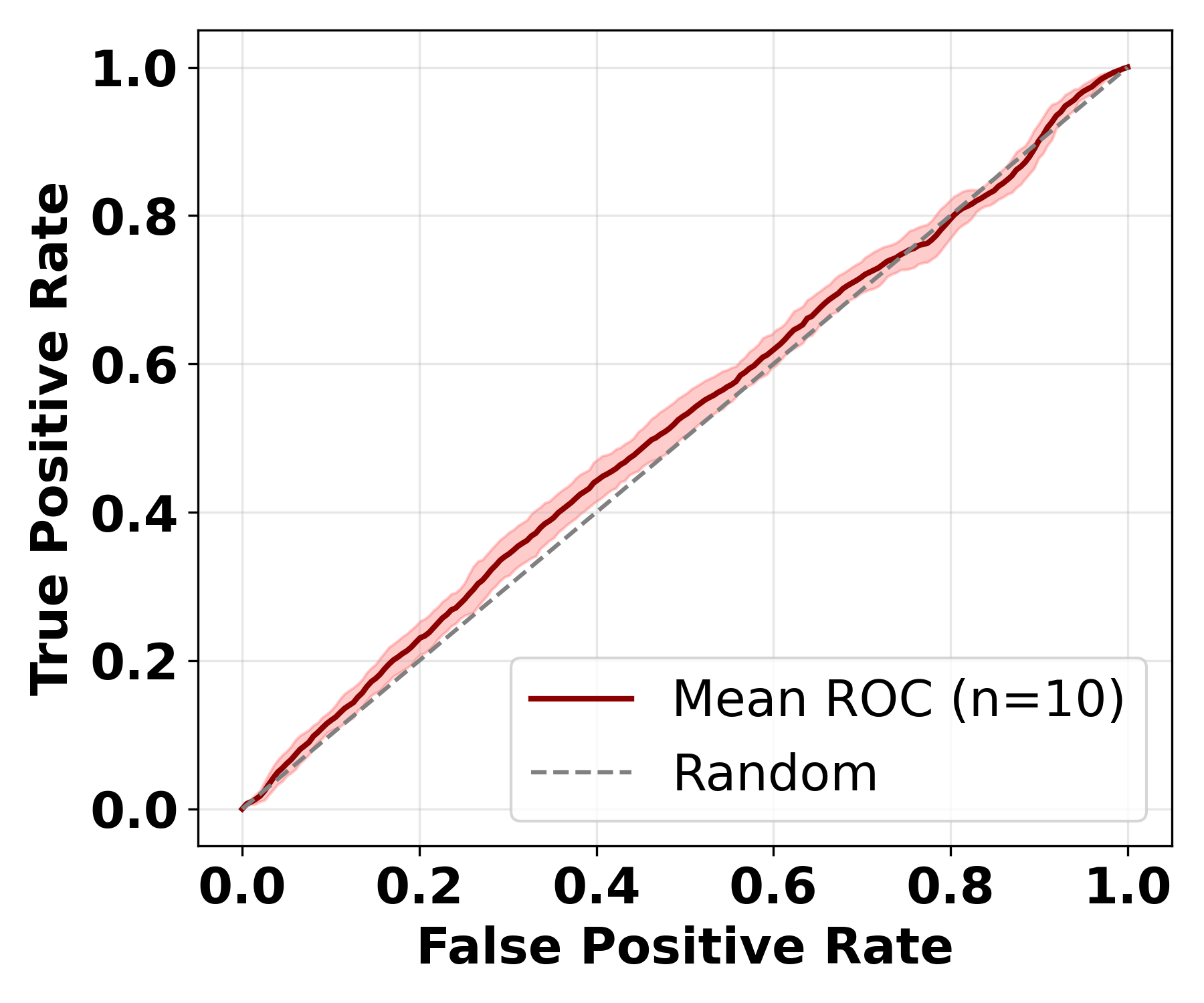}
        \caption{Grassland}
        \label{fig:TS_G_roc_curve}
    \end{subfigure}
    \caption{ROC curves with mean and standard deviation for forest and grassland regions under temporal split evaluation.}
    \label{fig:TS_ROC_curve_F_G}
\end{figure}

\begin{figure}[htbp]
    \centering
    % Forest PR Curve
    \begin{subfigure}[b]{0.49\linewidth}
        \centering
        \includegraphics[width=\linewidth]{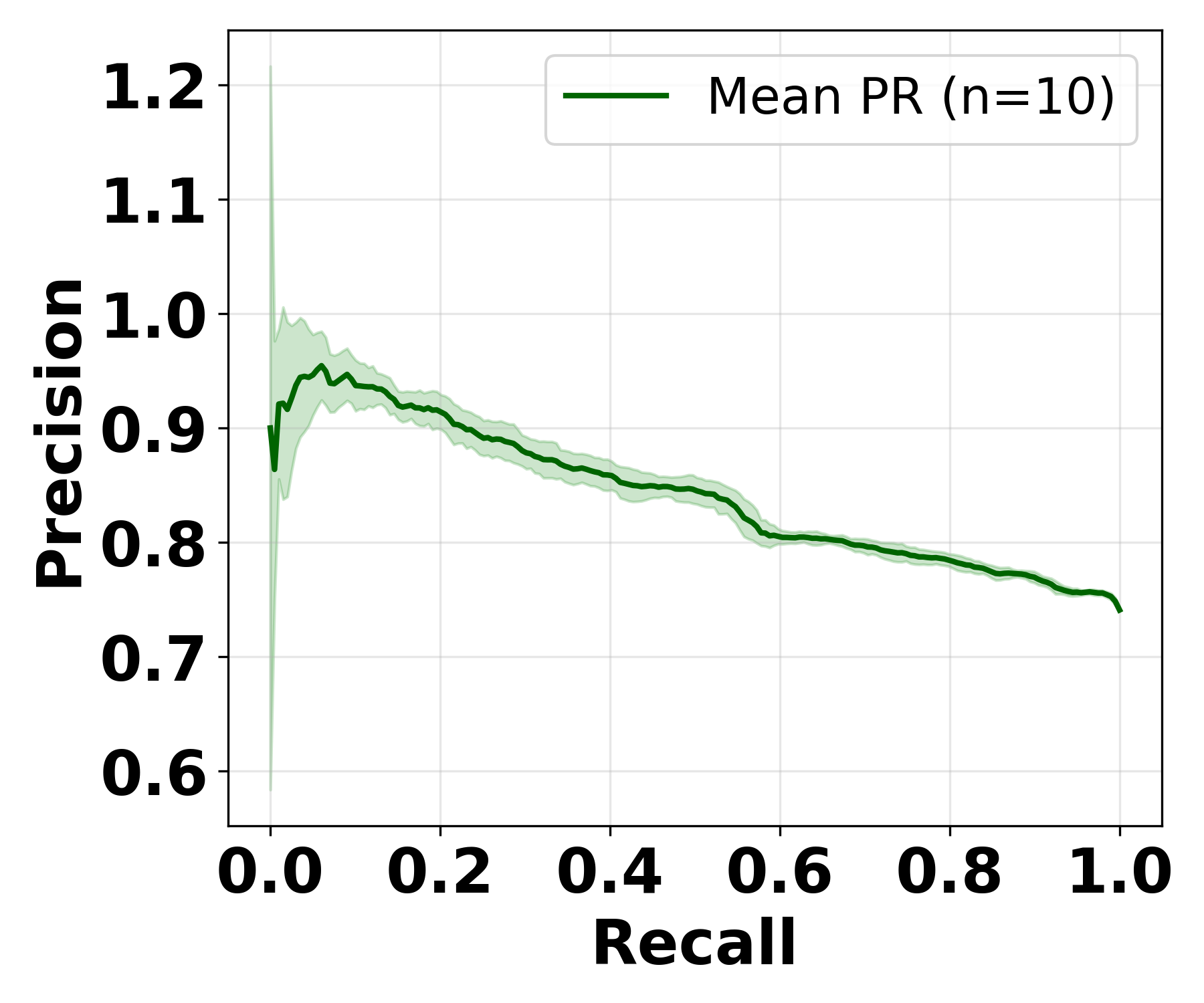}
        \caption{Forest}
        \label{fig:TS_F_pr_curve}
    \end{subfigure}
    \hfill
    % Grassland PR Curve
    \begin{subfigure}[b]{0.49\linewidth}
        \centering
        \includegraphics[width=\linewidth]{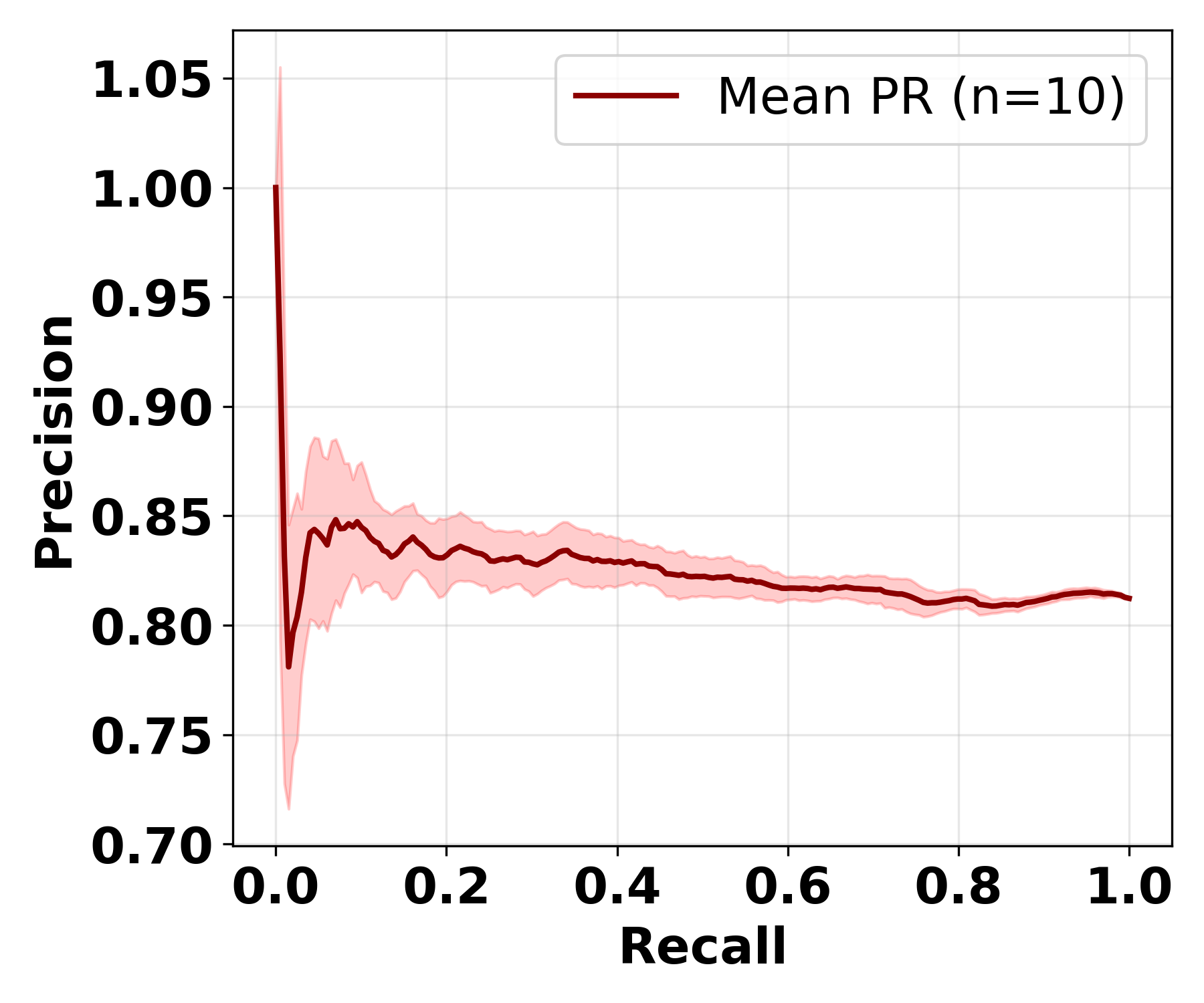}
        \caption{Grassland}
        \label{fig:TS_G_pr_curve}
    \end{subfigure}
    \caption{PR curves with mean and standard deviation for forest and grassland regions under temporal split evaluation.}
    \label{fig:TS_PR_curve_F_G}
\end{figure}

\begin{figure}[htbp]
    \centering
    % Forest Reliability
    \begin{subfigure}[b]{0.49\linewidth}
        \centering
        \includegraphics[width=\linewidth]{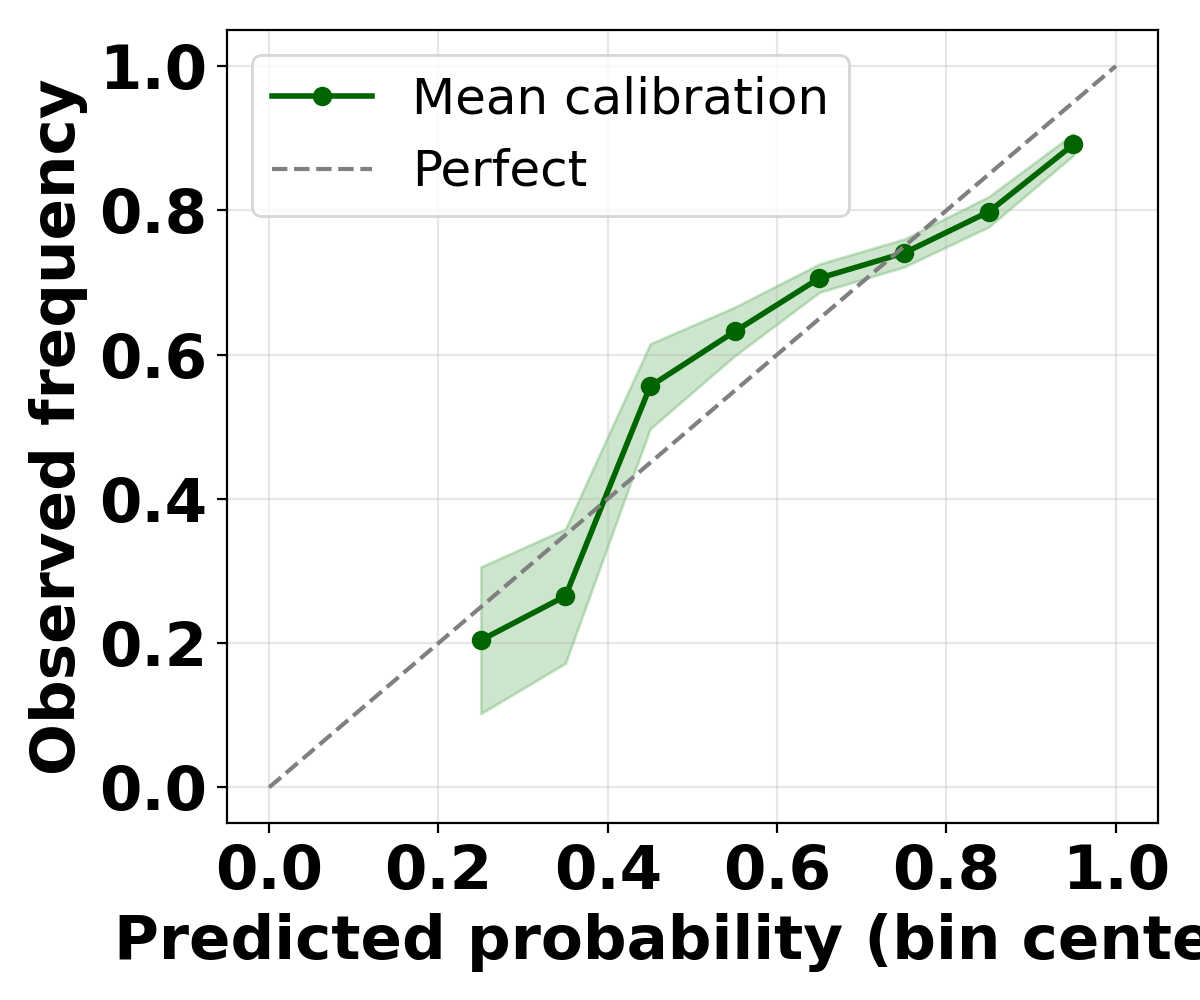}
        \caption{Forest}
        \label{fig:TS_F_reliability}
    \end{subfigure}
    \hfill
    % Grassland Reliability
    \begin{subfigure}[b]{0.49\linewidth}
        \centering
        \includegraphics[width=\linewidth]{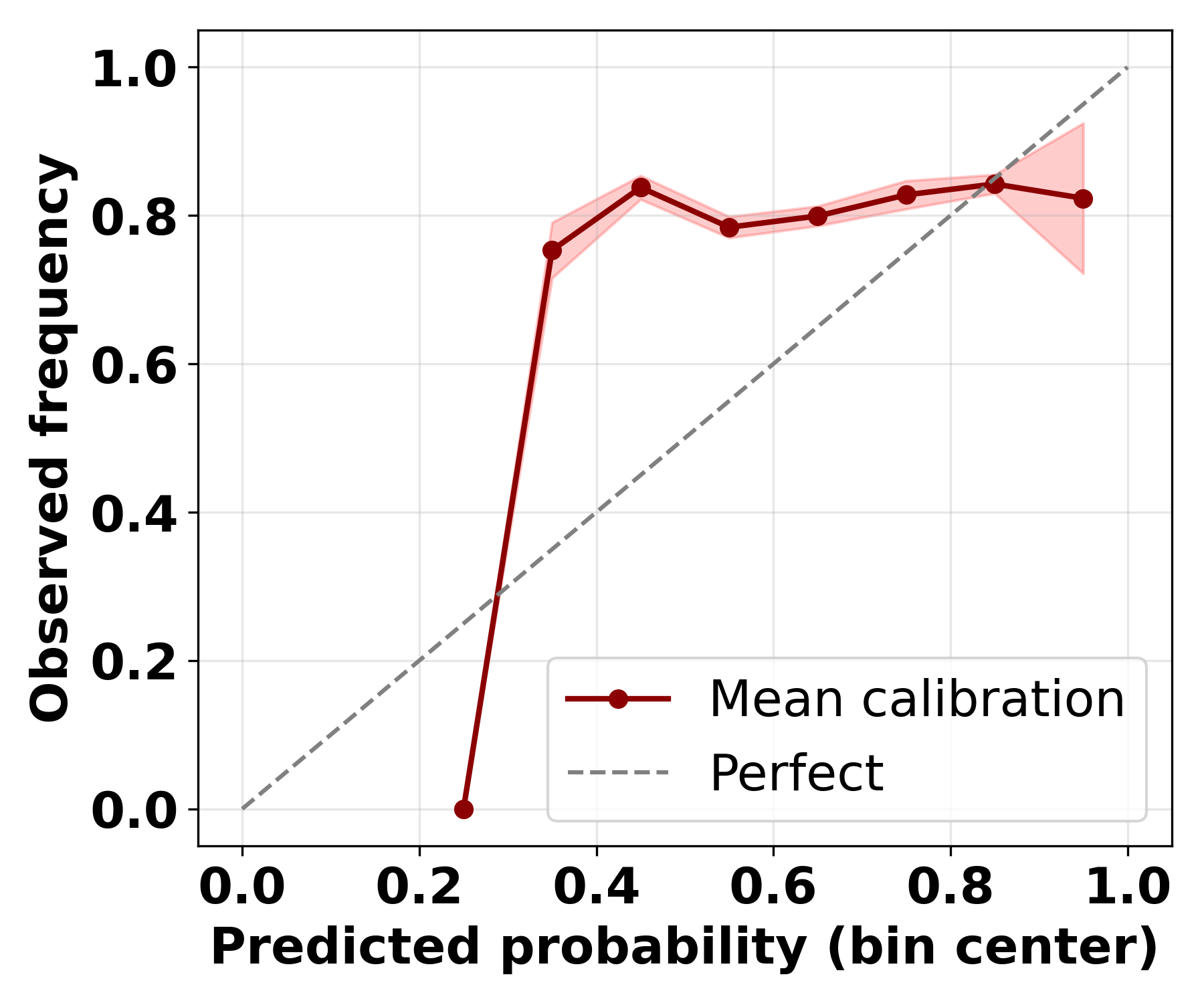}
        \caption{Grassland}
        \label{fig:TS_G_reliability}
    \end{subfigure}
    \caption{Reliability diagrams showing model calibration performance for forest and grassland regions under temporal split evaluation.}
    \label{fig:TS_Reliability_curve_F_G}
\end{figure}

%\subsubsection{Temporal Split Validation}
For temporal validation, the RF classifier (trained on 2024 and tested on 2025) indicates moderate discrimination but strong PR performance for the forest stratum, as noted above.    
 %Aggregate metrics across $K=10$ stratified bootstrap runs yield a mean ROC AUC of $\bar{A}_{\mathrm{ROC}} \approx 0.658 \pm 0.0106$ and a mean PR AUC of $\bar{A}_{\mathrm{PR}} \approx 0.842 \pm 0.0082$. 
 %At the operational threshold $t=0.5$ the classifier attains mean precision $\approx 0.756$ and very high recall $\approx 0.965$, producing a high $F_{1}$ score $\approx 0.848$. 
The high positive prevalence in the test set ($\text{positive\_rate}\approx 0.7394$) explains the combination of a moderate ROC-AUC with a high PR-AUC: when positives dominate, PR-based metrics emphasize correct identification of positives and remain elevated even if class separation across the full score range is only modest as shown in Fig. \ref{fig:TS_ROC_curve_F_G}, Fig. \ref{fig:TS_PR_curve_F_G}, and Fig. \ref{fig:TS_Reliability_curve_F_G}.

% ===================== Temporal Split - Forest vs Grassland =====================
% PR Curve

% ROC Curve

% Reliability Diagram

% Top-K Capture Curve
\begin{figure}[htbp]
    \centering
    % Forest Top-K
    \begin{subfigure}[b]{0.49\linewidth}
        \centering
        \includegraphics[width=\linewidth]{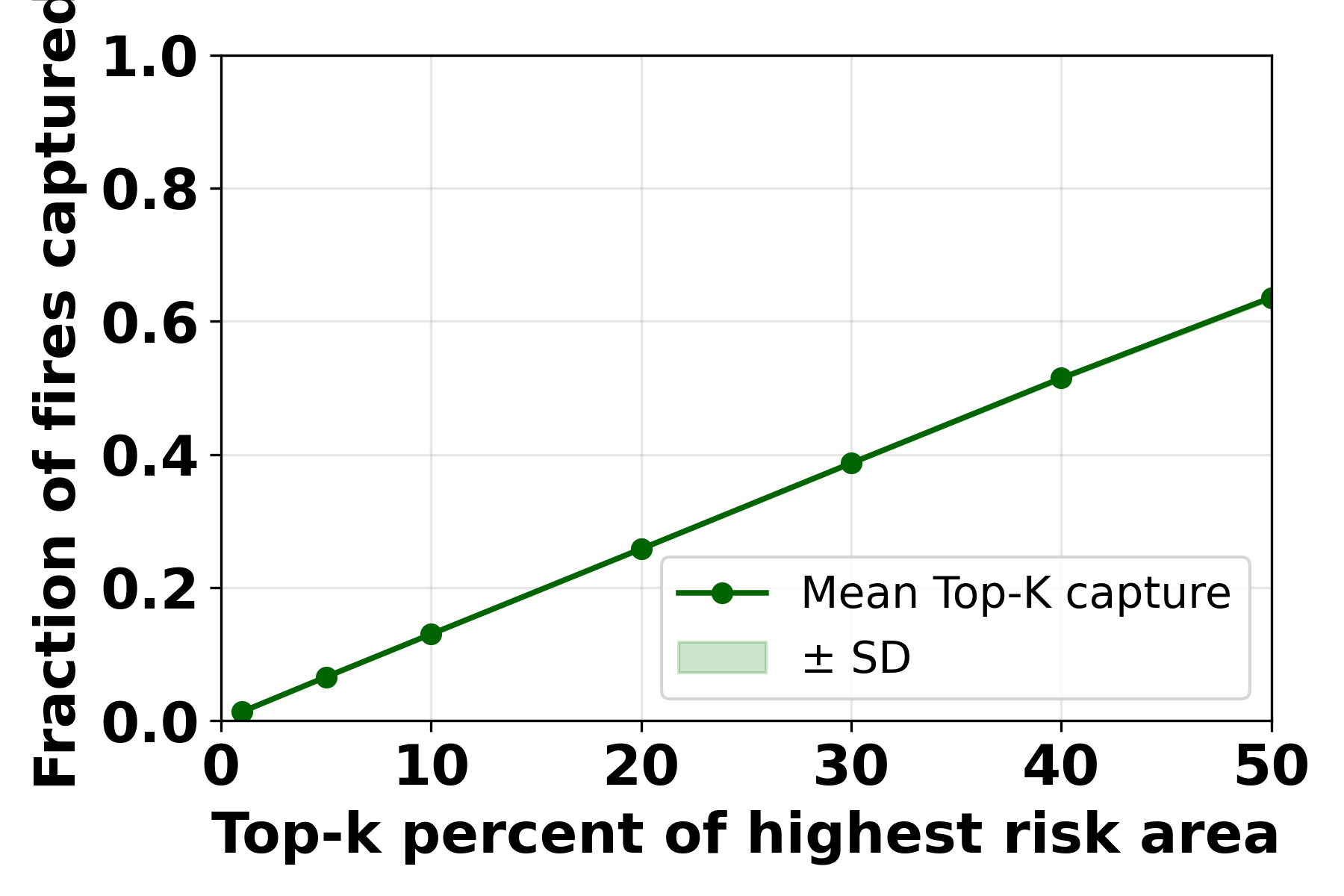}
        \caption{Forest}
        \label{fig:TS_F_topk_curve}
    \end{subfigure}
    \hfill
    % Grassland Top-K
    \begin{subfigure}[b]{0.49\linewidth}
        \centering
        \includegraphics[width=\linewidth]{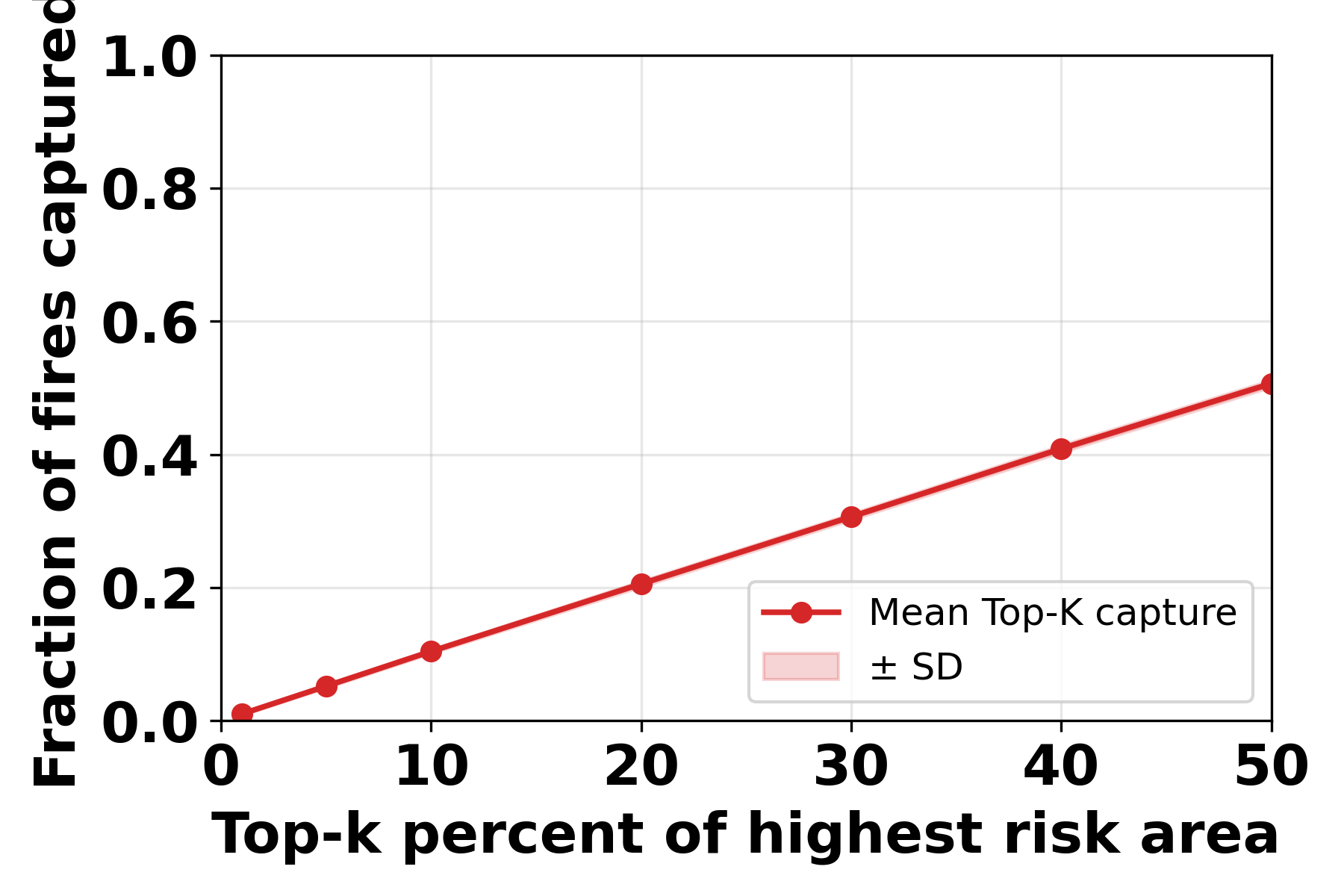}
        \caption{Grassland}
        \label{fig:TS_G_topk_curve}
    \end{subfigure}
    \caption{Top-$k$ capture curves for forest and grassland regions under temporal split evaluation.}
    \label{fig:TS_TopK_curve_F_G}
\end{figure}
Calibration and ranking diagnostics further qualified practical use. The mean Brier score ($\approx 0.182$) showed reasonable probabilistic accuracy but not perfect calibration, and the reliability diagram (mean $\pm$ SD) revealed systematic deviations in some probability bins (mid-range overconfidence and greater variability at extremes). Top-$k$ enrichment was modest: the top $50\%$ of ranked locations captured on average $\approx 55.7\%$ of forest positives (mean\_frac), so ranking improved prioritization but did not tightly concentrate true positives in the very top ranks as shown in Fig. \ref{fig:TS_TopK_curve_F_G}. 

\begin{table}[htbp]
\centering
\caption{Model performance under different validation types for Forest and Grassland ecosystems. The best performance metrics are highlighted in bold.}
\label{tab:model_performance_validation}
\resizebox{\linewidth}{!}{%
\begin{tabular}{l l c c c}
\hline
Ecosystem & \makecell{Split \\ Type} & ROC-AUC & PR-AUC & Brier Score \\ 
\hline
Forest & \makecell{Temporal Split \\ \textit{(Region-Transfer Test)}} & \textbf{0.6615} & \textbf{0.8423} & \textbf{0.1811} \\ 
Forest & \makecell{Spatial Split \\ \textit{(Training Performance)}} & 0.6155 & 0.8195 & 0.1892 \\ 
Grassland & \makecell{Temporal Split \\ \textit{(Region-Transfer Test)}} & 0.5235 & 0.8224 & 0.2025 \\ 
Grassland & \makecell{Spatial Split \\ \textit{(Training Performance)}} & 0.5416 & 0.8278 & 0.2117 \\ 
\hline
\end{tabular}%
}
\end{table}

Table~\ref{tab:model_performance_validation} summarizes model performance across validation types for both ecosystems. Forest ecosystems performed stronger, with the temporal split yielding the best metrics overall. 
 %ROC AUC $=0.6615$, PR AUC $=0.8423$, and Brier score $=0.1811$. Spatial split performance for forests is slightly lower, particularly in PR AUC ($0.8195$) and calibration ($0.1892$). For grasslands, performance is consistently modest in terms of ROC AUC ($\approx 0.52$--$0.54$) but relatively high in PR AUC ($\approx 0.82$), indicating limited overall class separability yet strong ability to prioritize true positives under imbalanced conditions. Brier scores ($\approx 0.20$--$0.21$) suggest moderate calibration.
This highlights that temporal transfer imposes stricter generalization demands and forests retain higher predictive separability and reliability than grasslands, likely reflecting stronger feature–class associations in forested regions.

\subsection{SHAP-Based Explainability}

%\subsubsection{\textbf{SHAP feature importance}}

% ===== SHAP Bar Plots =====
\begin{figure}[!ht]
    \centering
    \begin{subfigure}[b]{0.49\textwidth}
        \centering
        \includegraphics[width=\textwidth]{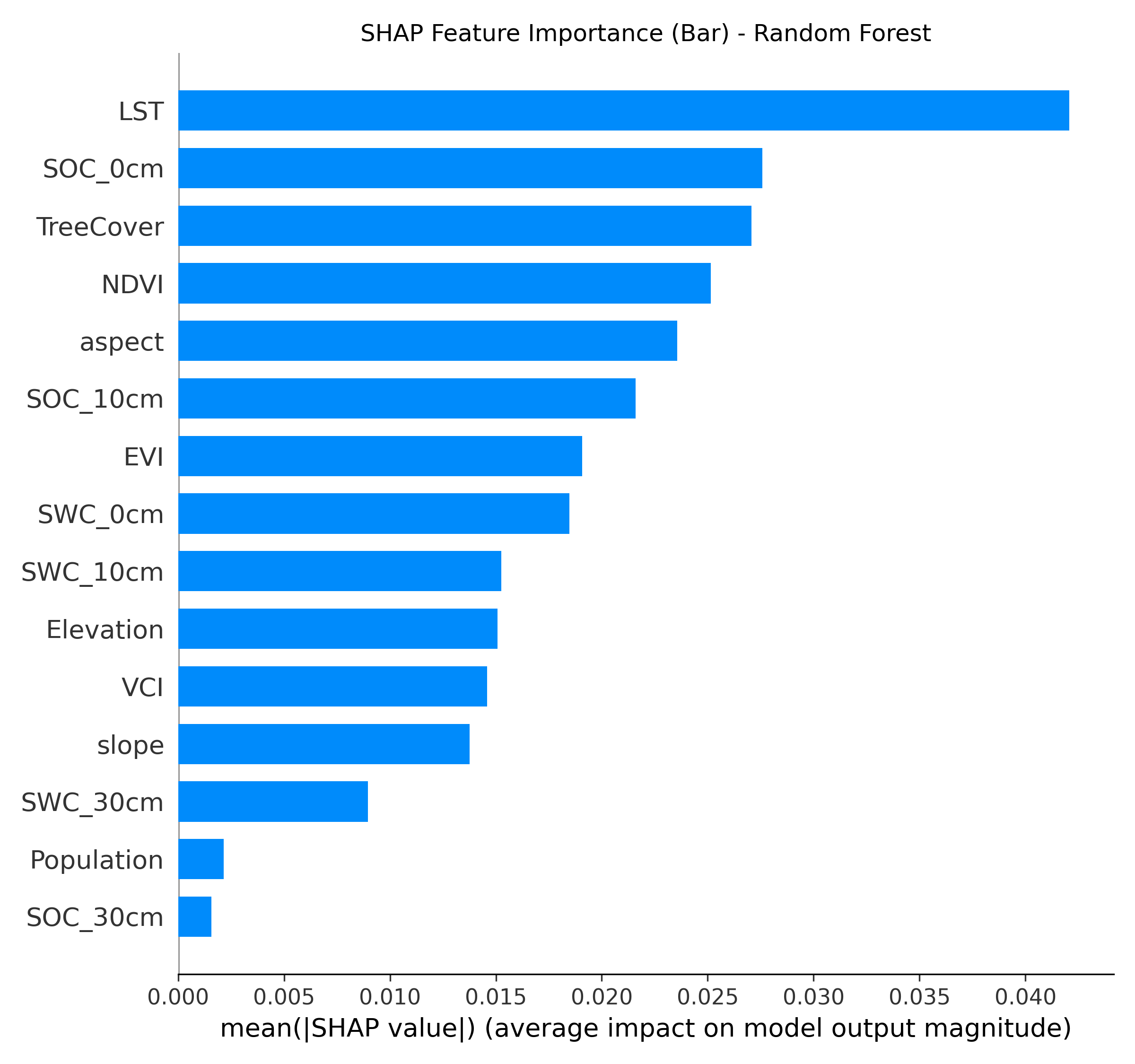}
        \caption{Forest}
        \label{fig:shap_bar_forest}
    \end{subfigure}
    \hfill
    \begin{subfigure}[b]{0.49\textwidth}
        \centering
        \includegraphics[width=\textwidth]{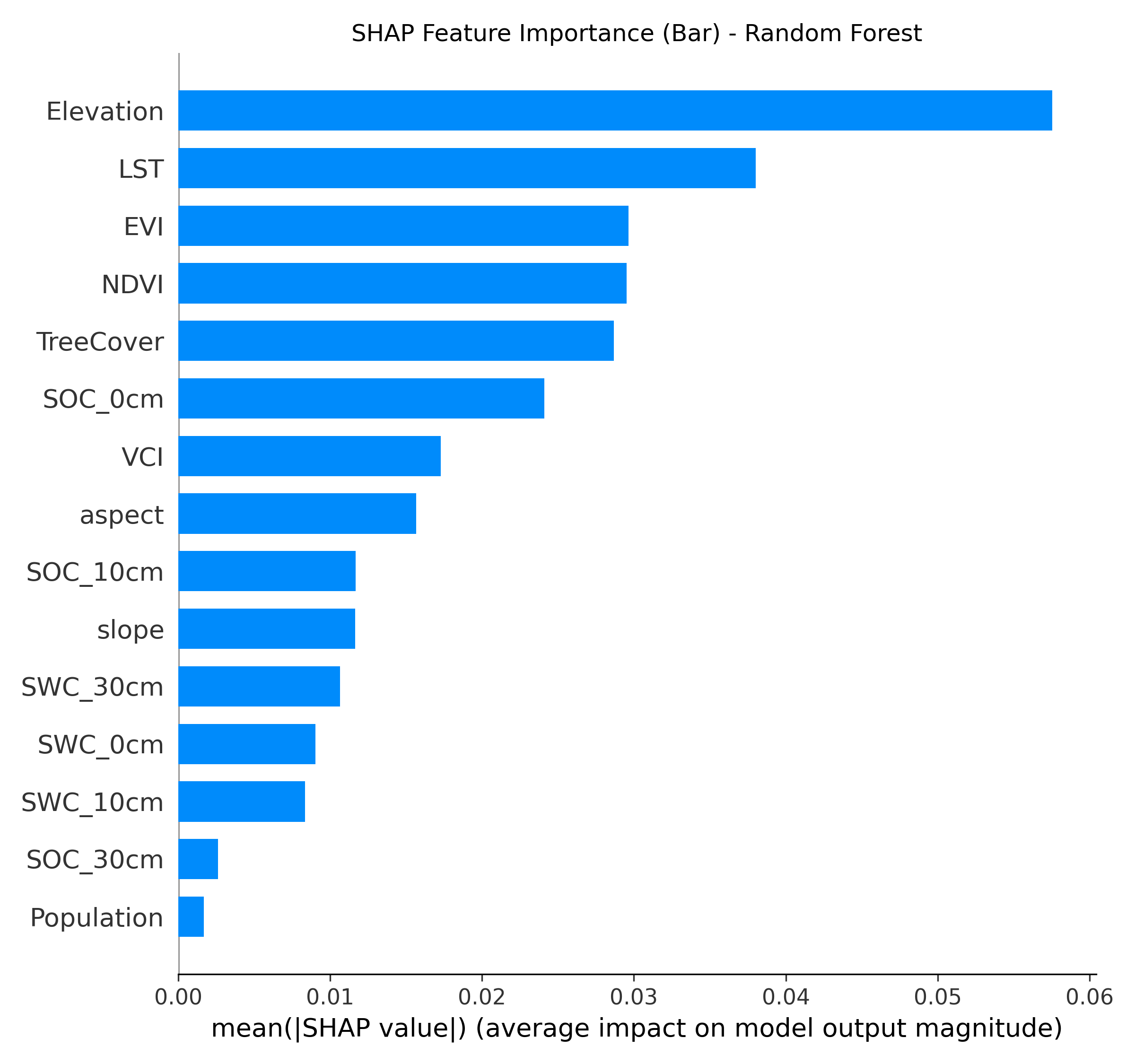}
        \caption{Grassland}
        \label{fig:shap_bar_grassland}
    \end{subfigure}
    \caption{SHAP feature importance bar plots for the RF classifier applied to forest and grassland datasets.}
    \label{fig:shap_barplots}
\end{figure}

% ===== SHAP Beeswarm Plots =====
\begin{figure}[!ht]
    \centering
    \begin{subfigure}[b]{0.49\textwidth}
        \centering
        \includegraphics[width=\textwidth]{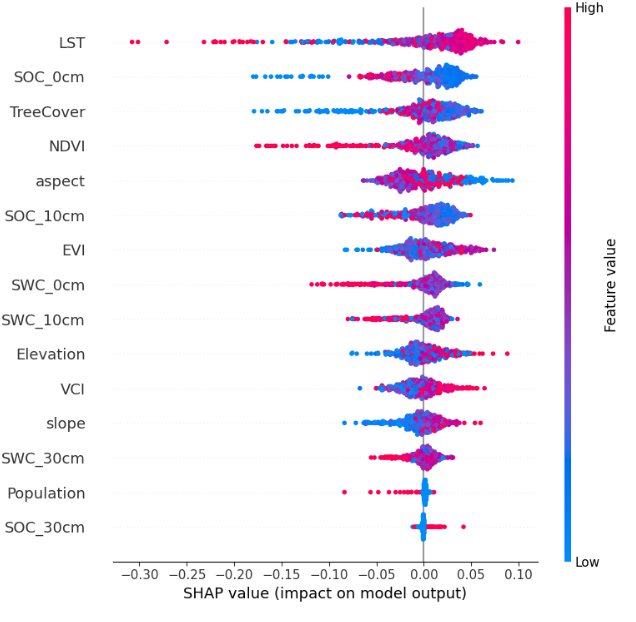}
        \caption{Forest}
        \label{fig:shap_beeswarm_forest}
    \end{subfigure}
    \hfill
    \begin{subfigure}[b]{0.49\textwidth}
        \centering
        \includegraphics[width=\textwidth]{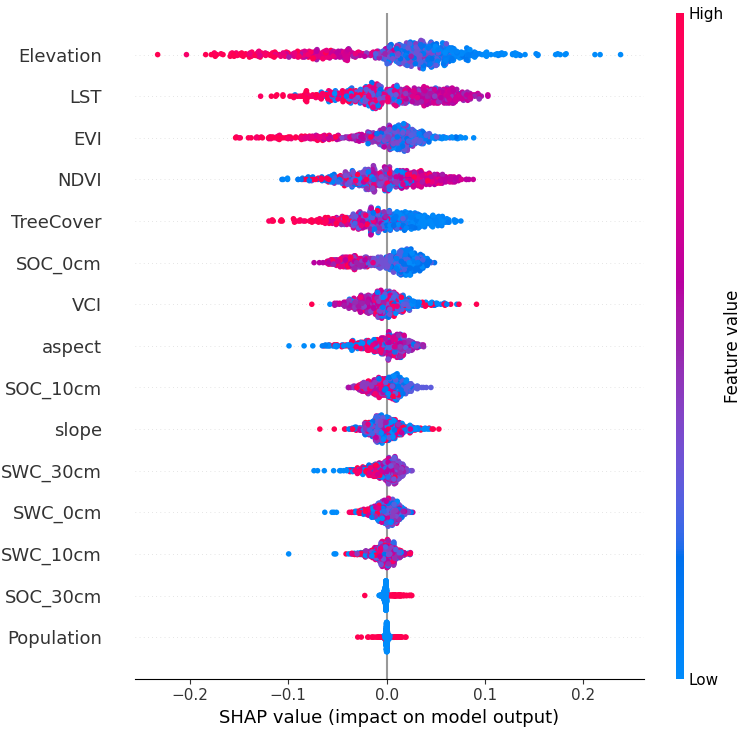}
        \caption{Grassland}
        \label{fig:shap_beeswarm_grassland}
    \end{subfigure}
    \caption{SHAP summary beeswarm plots for the RF classifier applied to forest and grassland datasets.}
    \label{fig:shap_beeswarmplots}
\end{figure}

The SHAP analysis of the RF model further revealed ecosystem-specific predictor importance as shown in Fig. \ref{fig:shap_barplots}. For forests, the most influential variables were soil organic carbon, tree cover, and NDVI, indicating that vegetation structure and soil properties strongly govern classification outcomes
 %Notably, higher values of elevation and LST were generally associated with positive contributions to the model’s predictions, underscoring their role in shaping grassland classification performance
in Fig. \ref{fig:shap_beeswarmplots}. Dense tree cover and steep slopes can facilitate fire spread in forests, while healthy vegetation, moist soils, and favorable aspects (e.g., north-facing slopes) reduce risk.

%\subsubsection{\textbf{\textbf{SHAP Force Plot}}}

\begin{figure}[!ht]
    \centering
    % --- Forest Force Plot ---
    \begin{subfigure}[b]{0.49\textwidth}
        \centering
        \includegraphics[width=\textwidth]{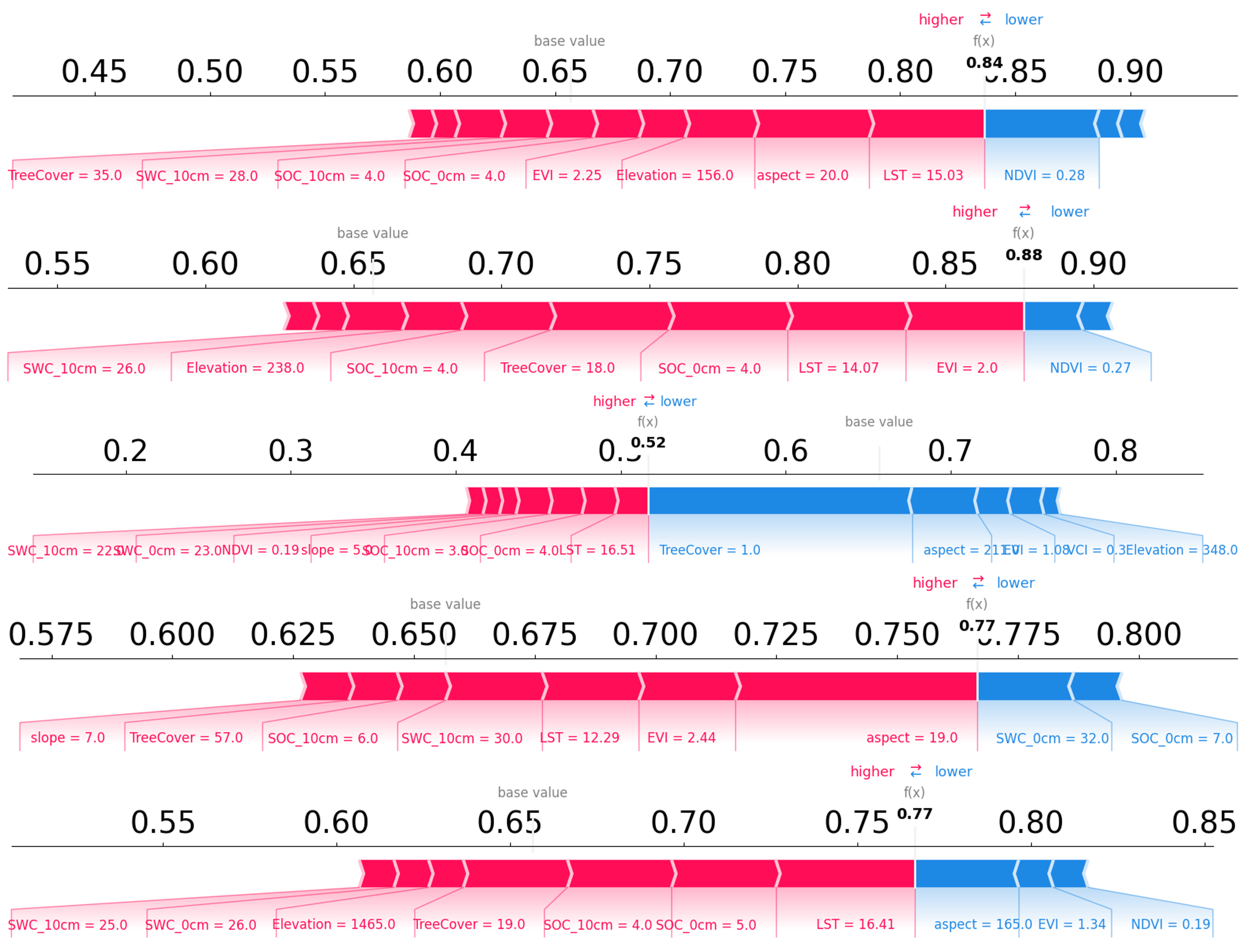}
        \caption{Forest}
        \label{fig:forceplot_forest}
    \end{subfigure}
    \hfill
    % --- Grassland Force Plot ---
    \begin{subfigure}[b]{0.49\textwidth}
        \centering
        \includegraphics[width=\textwidth]{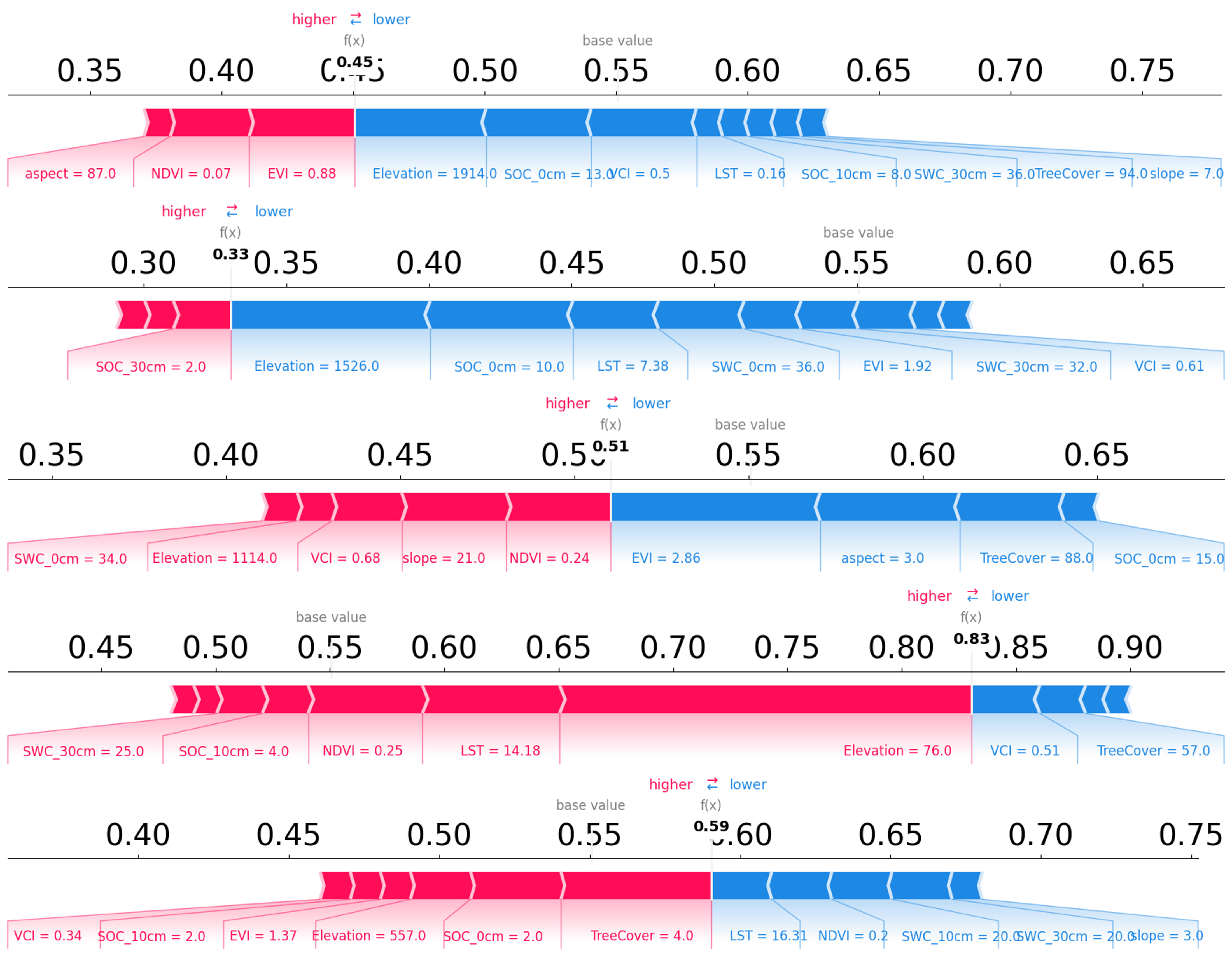}
        \caption{Grassland}
        \label{fig:forceplot_grassland}
    \end{subfigure}
    
    \caption{SHAP force plots illustrating the model's prediction explanation for forest and grassland ecosystems.}
    \label{fig:shap_force_plots}
\end{figure}

The SHAP force plots for grassland and forest fire predictions revealed key differences in how various environmental factors influence fire risk across land cover types as shown in Fig. \ref{fig:shap_force_plots}.    %Forest fire predictions show that while high temperature and low vegetation indices remain strong contributors to fire risk, additional factors such as tree cover, slope, and deeper soil moisture become     more influential. 
In contrast for grasslands, fire risk is primarily driven by high LST, low vegetation indices such as NDVI and EVI, and poor soil conditions — particularly low SOC and SWC at shallow depths. Elevation and aspect also play a role; higher elevations and certain slope orientations tend to reduce fire risk due to cooler and moister conditions.
\begin{figure}[!t]
    \centering
    
    % --- Map on top ---
    \includegraphics[width=\linewidth]{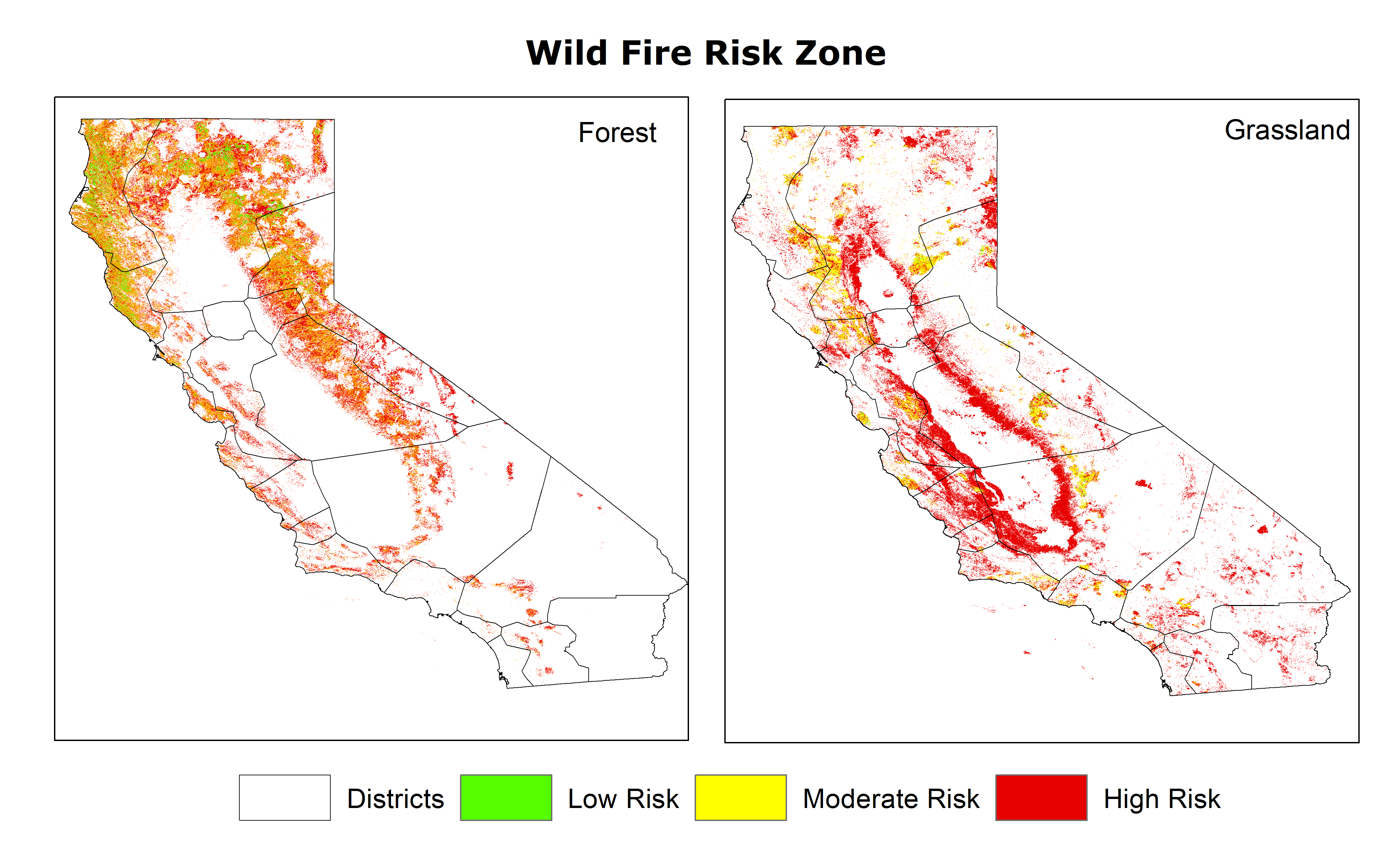}
    \vspace{0.5em} % small vertical space between map and graphs
    
    % --- Two graphs side-by-side ---
    \begin{subfigure}[b]{1\linewidth}
        \centering
        \includegraphics[width=\linewidth]{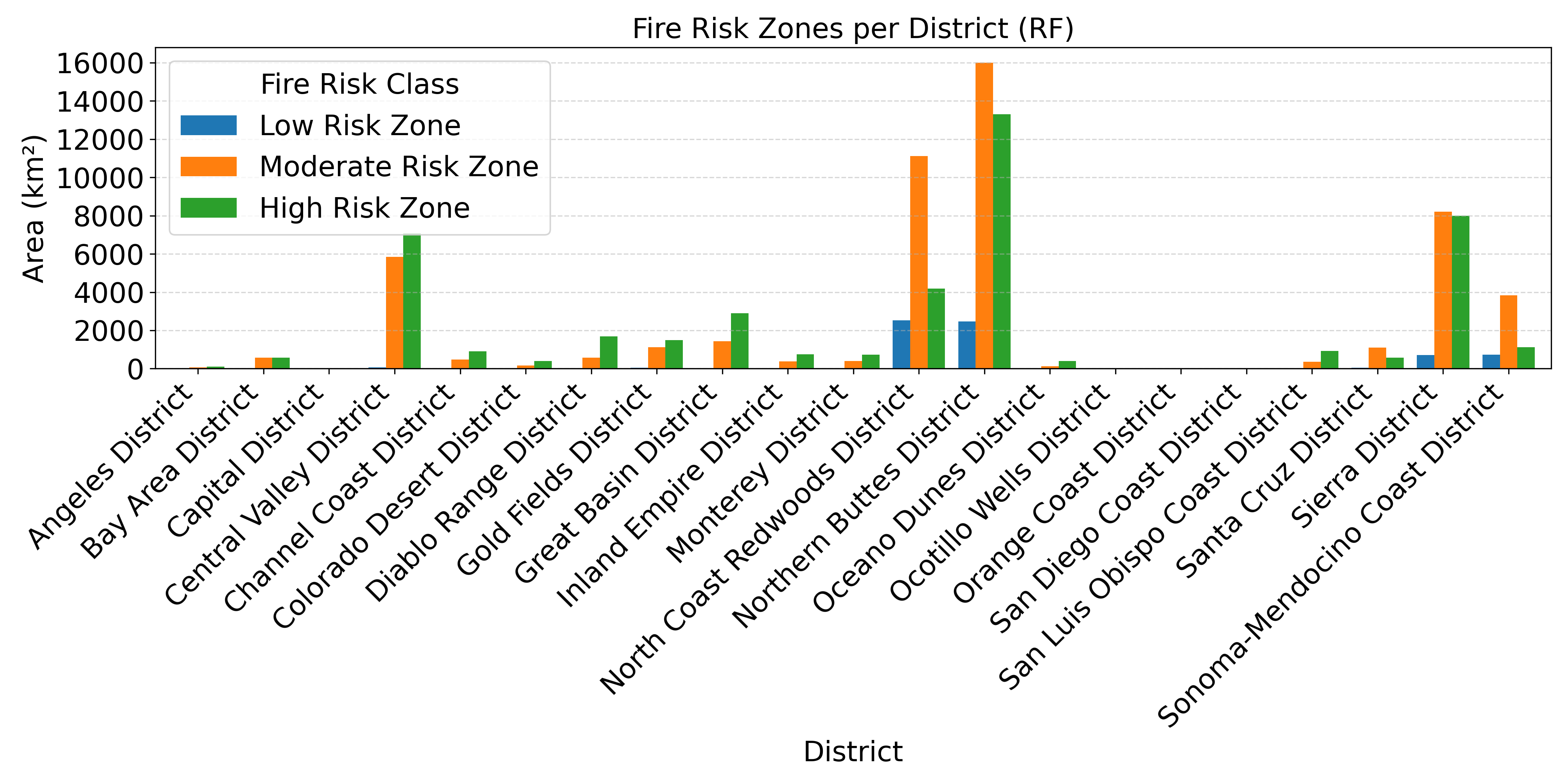}
        \caption{Forest}
        \label{fig:fire_risk_forest}
    \end{subfigure}
    \hfill
    \begin{subfigure}[b]{1\linewidth}
        \centering
        \includegraphics[width=\linewidth]{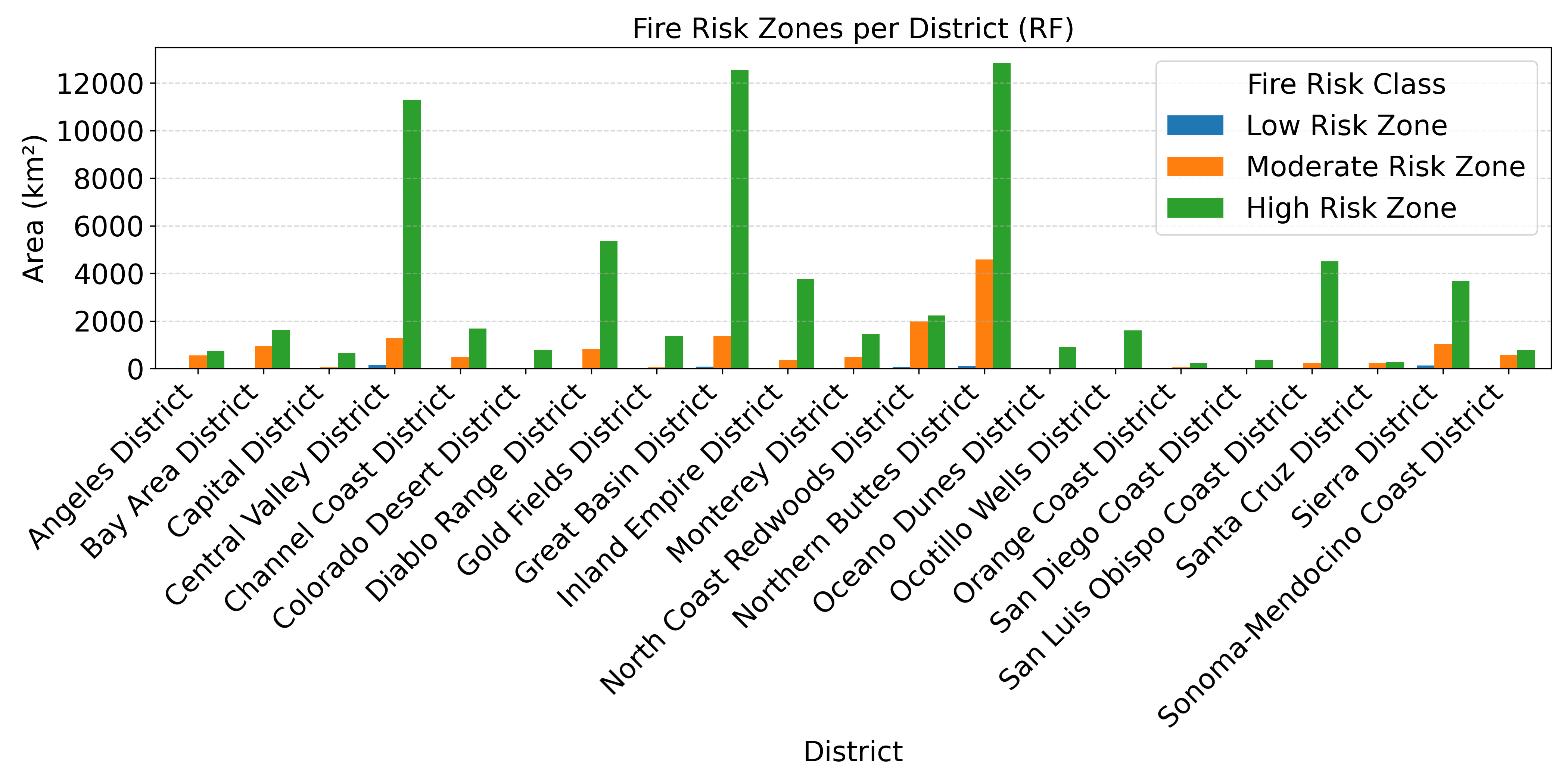}
        \caption{Grassland}
        \label{fig:fire_risk_grassland}
    \end{subfigure}
    
    % --- Main caption ---
    \caption{Fire risk assessment results: (top) spatial distribution of fire risk zones; (bottom) risk class distribution by district for forest and grassland ecosystems.}
    \label{fig:fire_risk_overview}
\end{figure}
\subsection{Fire Risk Zonation}
In forested areas, the Northern Buttes District emerged as the most dominant region, covering 31,781.41 km\textsuperscript{2}, with Class 2 (moderate) forests occupying 16,000.16 km\textsuperscript{2}, closely followed by Class 3 (high) forests at 13,311.30 km\textsuperscript{2} as in Fig. \ref{fig:fire_risk_forest}. The North Coast Redwoods District also exhibited significant forest cover (17,829.17 km\textsuperscript{2}), with a remarkable concentration of Class 2 forests (11,125.98 km\textsuperscript{2}), indicating dense and mature forest stands that are ecologically significant. In contrast, districts such as Capital District, Orange Coast, and San Diego Coast had minimal forest cover ($<$100 km\textsuperscript{2}), emphasizing their limited contribution to the overall forest extent. Interestingly, some regions, including the Channel Coast and Diablo Range Districts, demonstrated balanced distributions between Class 2 and Class 3 forests, reflecting transitional forest ecosystems.

The classification of grasslands across districts revealed a dominance of high-risk areas (Class 3) in terms of spatial coverage, indicating a higher susceptibility to disturbances such as fires or land degradation as shown in Fig. \ref{fig:fire_risk_overview}. The Central Valley District exhibited the largest extent of grassland, with 11,298.28 km\textsuperscript{2} falling under the high category, accounting for nearly 89\% of its total grassland area (12,718.13 km\textsuperscript{2}). Similarly, the Northern Buttes District showed a substantial proportion of Class 3 grasslands (12,848.23 km\textsuperscript{2}), reflecting intensive land cover pressures in this region as shown in Fig. \ref{fig:fire_risk_grassland}. In contrast, low-class grasslands (Class 1) were negligible across most districts, representing less than 1\% of the total grassland area, with notable occurrences in the Great Basin (76.81 km\textsuperscript{2}) and Northern Buttes (122.75 km\textsuperscript{2}) districts. Overall, grassland distribution patterns suggest that management strategies should prioritize Central Valley, Northern Buttes, and Great Basin.

\section{Discussion}

 %ML models are often referred as a black box, because of their complex structures that make it difficult to understand the internal decision-making process and how individual parameters influence the predictions. 
Numerous studies have applied ML algorithms to identify forest fire–prone areas; however, most have focused primarily on delineating susceptible zones, without addressing why such predictions occur in those specific areas -- a critical consideration for policy and decision makers. For instance, \cite{Abdollahi2023} identified elevation and NDMI as the most influential variables for forest fire occurrence, followed by rainfall and NDVI. Similarly, \cite{Tonbul2024} investigated forest fire susceptibility in the Mediterranean region and reported soil moisture as the most influential factor, followed by PDSI, LST, elevation, and other variables. In another study, \cite{Li2024} examined wildfire susceptibility across Europe, identifying land surface temperature (LST), solar radiation (SR), temperature condition index (TCI), normalized difference vegetation index (NDVI), precipitation, and soil moisture as the six most influential drivers, with mean absolute SHAP values of 0.135, 0.116, 0.049, 0.046, 0.039, and 0.033, respectively. Their analysis further showed that LST and SR were positively associated with wildfire risk, whereas TCI, NDVI, precipitation, and soil moisture demonstrated negative associations. Understanding such drivers and their tipping points is crucial for developing effective early warning systems. As these thresholds are approached, managers can increase vigilance, implement proactive measures, and strengthen monitoring to mitigate wildfire risks.

\section{Conclusion}
This study demonstrated the successful application of RF models integrated with XAI techniques to predict and interpret wildfire risk across California's diverse ecosystems. The RF model achieved outstanding performance, particularly for forested areas (AUC = 0.997, F1-score = 0.963) and grasslands (AUC = 0.996, F1-score = 0.968), indicating its strong predictive capabilities. The inclusion of spatial and temporal validation approaches revealed that forest ecosystems maintained higher predictive stability compared to grasslands, likely due to stronger correlations between environmental variables and fire occurrences in forested regions. SHAP analysis provided critical insights into the underlying drivers of fire risk, as mentioned previously. The district-level classification further highlighted high-risk regions such as the Northern Buttes and Central Valley Districts for both ecosystems, emphasizing the need for region-specific fire management strategies. These findings provide a valuable decision-support framework for prioritizing fire mitigation efforts and optimizing resource allocation. 

Looking ahead, this framework can be expanded to improve fire risk prediction and management under future climate and land-use change scenarios. Additionally, the inclusion of socio-economic factors, such as population density and human activity, could further refine risk mapping and support community-based fire management initiatives. Future research should also explore the use of deep learning models, such as U-Net or Transformer-based architectures, to capture complex spatio-temporal patterns in wildfire dynamics.

%By quantifying and visualizing feature contributions at both global and local scales, SHAP not only enhanced the interpretability of model outputs but also bridged the gap between predictive accuracy and actionable insights. This is particularly important for operational wildfire risk management, where decision-makers require both robust predictions and a clear understanding of the underlying drivers. The proposed RF–SHAP framework demonstrates that coupling high-performing machine learning algorithms with explainable AI can yield transparent, interpretable, and spatially explicit wildfire risk maps.
% Use IEEE bibliography style
\bibliographystyle{IEEEtran}
\bibliography{sample}   

\end{document}